\documentclass{article} % For LaTeX2e
\usepackage[preprint]{colm2025_conference}

\usepackage{algorithm}
\usepackage{algpseudocode}
\usepackage{microtype}
\usepackage{hyperref}
\usepackage{url}
\usepackage{booktabs}
\usepackage{amsmath}
\usepackage{lineno}
\usepackage{mathtools}
\usepackage{verbatim}
\usepackage{subcaption} 
\usepackage{svg}
\usepackage{multirow}
\usepackage{longtable}
\usepackage{arydshln}

\usepackage{wrapfig}
\usepackage{booktabs}

\newcommand\nnfootnote[1]{%
  \begin{NoHyper}
  \renewcommand\thefootnote{}\footnote{#1}%
  \addtocounter{footnote}{-1}%
  \end{NoHyper}
}

\definecolor{darkblue}{rgb}{0, 0, 0.5}
\hypersetup{colorlinks=true, citecolor=darkblue, linkcolor=darkblue, urlcolor=darkblue}

\title{Layered Unlearning for Adversarial Relearning}

\author{Timothy Qian, Vinith Suriyakumar, Ashia Wilson, Dylan Hadfield-Menell\\
MIT \\
\texttt{\{tcqian,vinithms,ashia07,dylanhm\}@mit.edu} 
}

\begin{document}

\raggedbottom
\ifcolmsubmission
\linenumbers

\fi

\maketitle
\begin{abstract}
Our goal is to understand how post-training methods, such as fine-tuning, alignment, and unlearning, modify language model behavior and representations. We are particularly interested in the brittle nature of these modifications that makes them easy to bypass through prompt engineering or relearning. Recent results suggest that post-training induces shallow context-dependent ``circuits'' that suppress specific response patterns. This could be one explanation for the brittleness of post-training. To test this hypothesis, we design an unlearning algorithm, Layered Unlearning (LU), that creates distinct inhibitory mechanisms for a growing subset of the data. By unlearning the first $i$ folds while retaining the remaining $k - i$ at the $i$th of $k$ stages, LU limits the ability of relearning on a subset of data to recover the full dataset. We evaluate LU through a combination of synthetic and large language model (LLM) experiments. We find that LU improves robustness to adversarial relearning for several different unlearning methods. Our results contribute to the state-of-the-art of machine unlearning and provide insight into the effect of post-training updates.
\end{abstract}

\nnfootnote{Synthetic experiment code at: \href{https://github.com/12tqian/layered-unlearning}{\texttt{https://github.com/12tqian/layered-unlearning}}}

\section{Introduction}
Post-training interventions such as fine-tuning, preference learning, and unlearning are widely used to modify the behavior of pre-trained large language models (LLMs). However, changes introduced in post-training are often brittle. However, these changes are often shallow or brittle. In many cases, they are bypassed or reversed by clever adversarial prompting or fine-tuning~\citep{jain2024makesbreakssafetyfinetuning,arditi2024refusal,zou2023universal,greenblatt2024stresstesting,che2024model,deeb2025unlearningmethodsremoveinformation,betley2025emergentmisalignmentnarrowfinetuning}. Our goal is to understand how post-training methods modify language model behavior and representation and support the design of more robust post-training methods.

We study this through the lens of machine unlearning, which seeks to remove knowledge or capabilities from pre-trained models. \citet{deeb2025unlearningmethodsremoveinformation} recently demonstrated that ``unlearned'' information is easily re-elicited by fine-tuning on a subset of the removed data. To explain this result, we hypothesize that SoTA unlearning methods introduce a context-dependent \emph{inhibitor} mechanism. Efficient ``relearning'' generalizes because fine-tuning removes a single shared mechanism and reverses the full post-training modification. 

\begin{figure}[t]
\begin{center}
\includegraphics[width=\textwidth,scale=1]{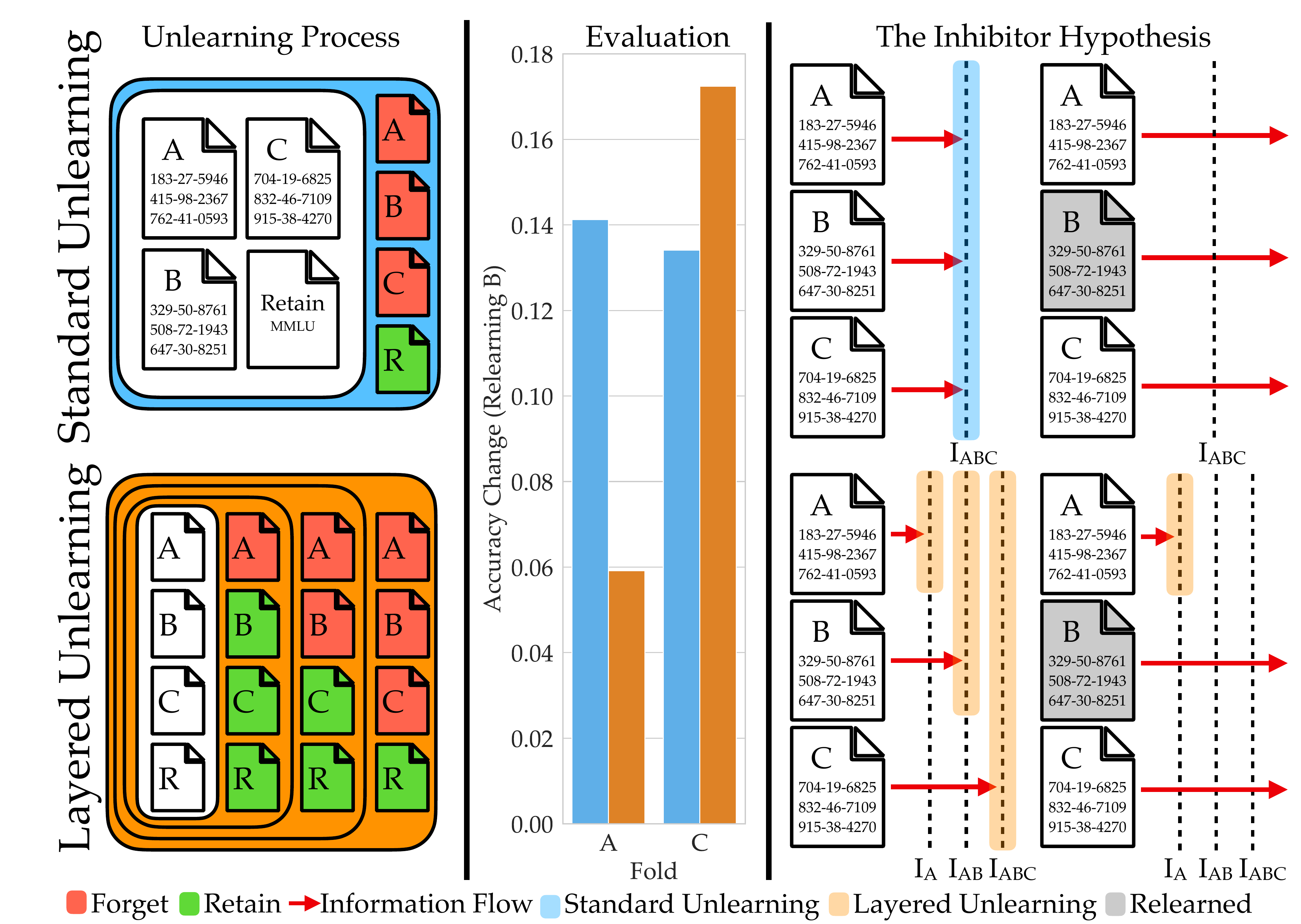}
\end{center}
\caption{
\textbf{Left:} An illustration of LU with social security numbers (SSNs). The SSNs are partitioned into disjoint sets $(A, B, C)$. \textit{Top:} Standard unlearning minimizes performance on $A\cup B \cup C$ while retaining general capabilities on a retain set $R$ (e.g., MMLU). \textit{Bottom:} In LU, we sequentially unlearn the sequence $\{A, A\cup B, A\cup B \cup C\}$ while retaining the sequence $\{B \cup C \cup R, C \cup R, R\}$. 
\textbf{Middle:} As a result, relearning $B$ improves performance on $C$ but not $A$. In contrast, training on any subset improves performance across the board for standard methods. \textbf{Right:} We hypothesize that unlearning the full set introduces a context-dependent shared \emph{inhibitor}  $I_{ABC}$ that suppresses the information and that subsequent relearning removes $I_{ABC}$. The structure of LU is designed to create several distinct inhibitors $I_{A}, I_{AB}, I_{ABC}$ that cover different folds of the data. Relearning on $B$ removes $I_{AB}$ and $I_{ABC}$, but leaves $I_{A}$ active. }
\label{fig:process}
\end{figure}

A common way to mitigate single failure points is the famous ``Swiss cheese'' defense-in-depth model~\citep{reason1990human}. The goal is to combine multiple imperfect defenses. If their failure modes are distinct, the combined defense is more robust than any individual approach. We implement defense-in-depth through \textbf{Layered Unlearning} (LU). LU partitions the data into $k$ disjoint folds and applies unlearning sequentially to a growing subset: at stage $i$, LU unlearns the union of folds $F_1$ through $F_i$. Crucially, we retain the data from $F_{i+1}$ through $F_k$ to induce distinct inhibitors at each stage of unlearning. Figure~\ref{fig:process} illustrates the algorithm and the defense-in-depth inspiration.

We investigate the performance of LU on a variety of synthetic tasks and unlearning benchmarks. We consider a synthetic 2-dimensional classification task and a 3-token sequence generation task. Next, we apply LU to a variety of unlearning methods on the WMDP, Years, and MMLU datasets. In all settings, LU improves resistance to fine-tuning-based recovery. In the course of these experiments, we also identify a stronger class of attack: \emph{corpus-based fine-tuning}. This attack, which uses raw text rather than structured MCQ-based prompts, bypasses inhibitors more effectively than standard RTT~\citep{deeb2025unlearningmethodsremoveinformation}. Notably, this distinction only emerges because LU creates variation in robustness across data folds—an effect that is not observed with standard unlearning.

We make three contributions: 1) we introduce \emph{Layered Unlearning} (LU) a method that combines multiple steps of unlearning to increase robustness; 2) we show in both synthetic and LLM settings that LU improves resistance to adversarial relearning; and 3) we use LU to reveal a gap in attack strength between MCQ-based and corpus-based relearning, offering new insight into the limits of post-training behavioral control. Our results contribute to the state-of-the-art of machine unlearning and provide insight into the effect of post-training updates.

\section{Layered Unlearning}

In this section, we introduce LU. We begin with a replication of \citet{deeb2025unlearningmethodsremoveinformation} in two synthetic tasks: a 2D classification task with mixtures of Gaussians and a bigram completion task with three tokens. Next, we introduce the LU algorithm. Finally, we investigate the effect of LU in this task. We find that LU improves robustness in both cases and analyze the sequences of changes that LU induces. First we introduce some notation to represent a machine unlearning method $U$. 

\subsection{Machine unlearning notation} The goal of machine unlearning is to remove information $F$ from trained model weights $\theta \in \Theta$ that model a dataset $D$ in some input space $X$. With unlimited compute, this would involve retraining from scratch on $  D\setminus F$. Due to, e.g.,  the cost of pretraining, unlearning methods attempt to approximate this result as a post-training step that maintains performance on a retain set $R\subset D$.

Thus, we can represent a generic unlearning algorithm $U$ as a function that maps model parameters $\theta$, forget set $F$, retain set $R$, and hyperparameters $\gamma \in \Gamma$ to a new set of model parameters $\theta'$. When clear from context, we may omit the final argument corresponding to the hyperparameters. Formally 

\[U:\Theta \times X \times X \times \Gamma \rightarrow \Theta.\]

\subsection{Adversarial relearning in synthetic settings}

We replicate the results of \citet{deeb2025unlearningmethodsremoveinformation} in two synthetic settings: a 2D classification task and a bigram language modeling task.

\paragraph{2D logistic regression.} Our first task is a 2D logistic regression task, so our input space is $X = \mathbb{R}^2$. The goal is to classify a mixture of Gaussians (class $1$) against a uniform background distribution over $\mathcal  U([-60, 60]^2)$ (class $0$). We sample the Gaussian means from the uniform distribution $\mathcal U([-50, 50]^2)$ and use a primarily isotropic covariance matrix with variance $\sigma^2 = 4$, adding small perturbations of magnitude $0.1$ to break exact symmetry. We implement a linear classifier with logistic regression on radial basis functions (RBF) features. 

\begin{figure}[t]
    \centering
    \includesvg[width=\textwidth]{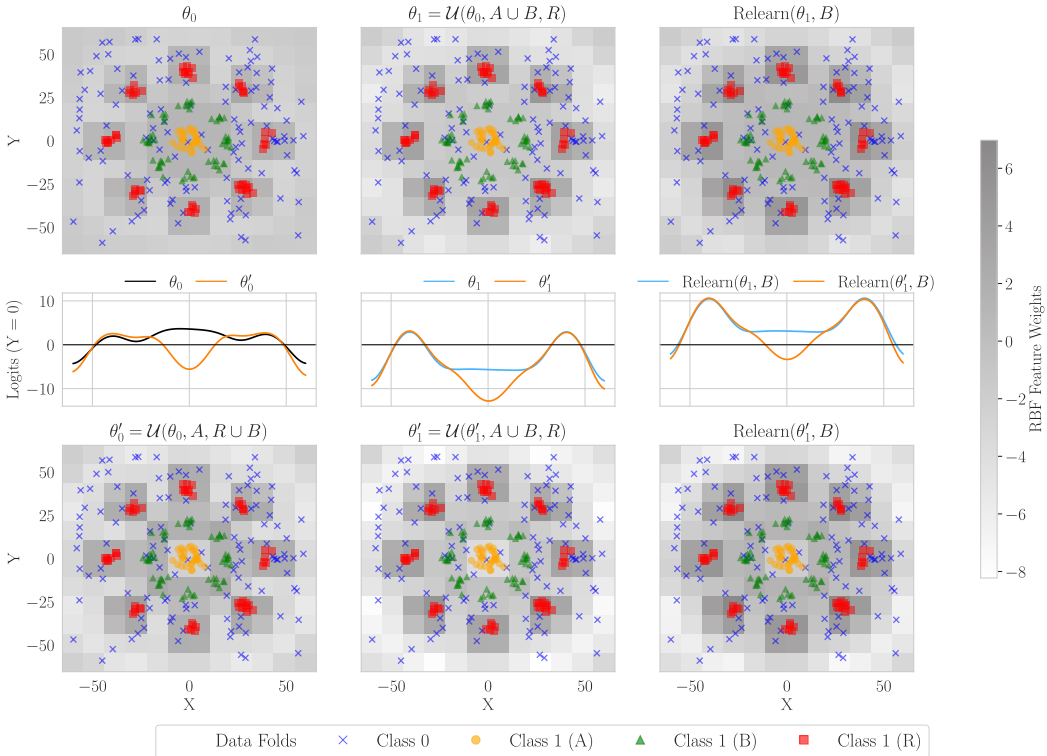}
    \caption{
       A depiction of Layered Unlearning in our 2D logistic regression example. Scatter plots represent the data, which consists of a uniform distribution (Class 0) and a mixture of Gaussians (Class 1) that is split into three subsets, $A$, $B$, and $R$. The goal of unlearning is to forget $A, B$ while retaining $R$. Our classifiers are trained with logistic regression with radial basis functions spaced out in a grid. We show the weights as a heatmap on the grid. The top left shows $\theta_0$, the initial trained model. Across the top row, we illustrate the effect of joint unlearning $\theta_1 = (\theta_0, A \cup B, R)$  and subsequent relearning on $B$. Note that relearning $B$ also relearns $A$. The corresponding classification logits along $Y=0$ are shown below. Notice how learning on $B$ generalizes to the area around $(0, 0)$. In the bottom row, we show the steps of Layered Unlearning. First, we compute $\theta'_0 = U(\theta_0, A, R\cup B),$ shown in the bottom left, then we compute $\theta_1' = U(\theta_0', A \cup B, R),$ shown in the bottom middle. The logit plot shows the clear effect on the logits near $(0, 0)$. The bottom right shows the effect of subsequently relearning $B$, while performance on $B$ still improves, it no longer generalizes to $A$.  
       %Visualization of inhibitors. The initial model $\theta_0$ classifies all Gaussian mixtures as class $1$ and the Null set as class $0$. Task $A$ (resp. $B$) unlearning reclassifies mixture $A$ (resp. $B$) as class $0$. The retain task evaluates accuracy on the retain mixture (class $1$) and Null (class $0$). We use logistic regression with radial basis function (RBF) features centered on a $12 \times 12$ grid. Each panel shows Gaussian scatters and a heatmap of RBF weights (not logits or probabilities). Standard unlearning ($\theta_1$) learns a broad inhibitor (negative-weight disk), while Layered Unlearning ($\theta_1'$) learns $I_A$ and then $I_{AB}$—a localized inhibitor and a broad inhibitor in sequence. Upon relearning $B$, standard unlearning ($\mathrm{Relearn}(\theta_1, B)$) collapses, recovering $A$, while LU $\mathrm{Relearn}(\theta_1', B)$) preserves $I_A$ and continues suppressing $A$. The center row plots logits along $y = 0$, highlighting the spatial extent of each inhibitor.
    }
    \label{fig:gmm_heatmap}
\end{figure}

We partition the Gaussians into subsets $A, B, R$, where the goal is to unlearn $A \cup B$ and retain $R$. Tasks $A$ and $B$ are defined as the classification accuracy when sets $A$ and $B$ are labeled as class $1$, and the retain task as the joint accuracy on classifying $R$ as class $1$ and Null as class $0$. We place the RBF centers on an $12\times12$ grid of points, so $\Theta = \mathbb{R}^{145}$ (including a bias term). The top left of figure~\ref{fig:gmm_heatmap} shows the learned weights with $2$ folds.

The unlearning primitive $U$ takes as input model parameters $\theta$, a forget set $F$, a retain set $R$, and hyperparameters $\gamma$. Each data point in $F$ and $R$ is a 2D input. The objective is to preserve the original classification for points in $R$, while reclassifying points in $F$ as class $0$. Relearning refers to assigning data points in the relearned set back to their original classifications. We optimize all objectives using the Adam optimizer.

\paragraph{Bigram sequence modeling.} Next, we consider a bigram language modeling task with three tokens: $a, b, r$. The input space is length $8$ token sequences: $X = \{a, b, r\}^8$. We generate data so that $a$ and $b$ are followed by $r$, while $r$ is followed by $a$ and $b$ with equal probability. We combine this with a small minimum probability $\epsilon = 0.05$ of a uniform transition across tokens to encourage smooth learning dynamics. This leads to the following conditional probabilities for consecutive tokens: $P(r\mid a)=P(r\mid b)=1-2\epsilon$, $P(a\mid r)=P(b\mid r)=\frac{1}{2}-\frac{\epsilon}{2},$ and $P(a\mid a)=P(b\mid a)=P(a\mid b)=P(b\mid b)=P(r\mid r)=\epsilon.$ 

We define tasks $A$, $B$, and $R$ as the prediction performance on all consecutive pairs tokens in the bigram sets $\{aa, ab, ar\}$, $\{ba, bb, br\}$, and $\{ra, rb, rr\}$, respectively. We use a small attention-only, one-layer transformer with parameter space $\Theta = \mathbb{R}^{4288}$. This architecture is simple enough to permit analytical study~\citep{elhage2021mathematical}, while still serving as a useful proxy for the larger models used in our later experiments.

The unlearning primitive $U$ takes as input model parameters $\theta$, a forget set $F$, a retain set $R$, and hyperparameters $\gamma$. Data points in $F$ and $R$ are elements of $\{a, b, r\}$. For each $x \in R$, the objective is to preserve the original conditional distribution over consecutive tokens, $P(\cdot \mid x)$. For each $x \in F$, the goal is to flatten the distribution to $P(\cdot \mid x) = \frac{1}{3}$. The retain task prevents the model from collapsing to a trivial uniform distribution over tokens after unlearning. To implement unlearning, we generate sequences from this modified transition matrix and optimize the standard language modeling loss. Relearning refers to restoring the original conditional distribution over tokens for the subset being relearned. We optimize all objectives using the Adam optimizer.

% report averaged across 10 seeds
\textbf{Relearning performance.} For each task, we first apply the unlearning primitive, followed by relearning on each subtask. We limit our discussion to relearning on $B$ and evaluating on $A$ for symmetry's sake. In 2D classification, relearning on $B$ restores $93\%$ of task $A$'s original performance (Table~\ref{tab:gmm_acc}). Figure~\ref{fig:gmm_heatmap} (top) illustrates the corresponding weight trajectories. In bigram modeling, we observe similar behavior: relearning on $B$ recovers $73\%$ of task $A$'s performance (Table~\ref{tab:bm_acc}).

\subsection{The Layered Unlearning algorithm}

Next, we present the LU algorithm and evaluate it in these two domains. Algorithm~\ref{alg:seq_unlearning} shows the algorithm details. The algorithm relies on an unlearning primitive $U$ that maps model weights $\theta$, forget set $F$, retain set $R$, and algorithm hyperpararmeters $\gamma$ 

Its primary input is a set of model weights $\theta_0$, a sequence of $k$ forget sets $\{F_i\}$, a retain set $R$, an unlearning algorithm $U$, and a sequence of algorithm hyperparameters $\{\gamma_i\}$. LU proceeds through $k$ steps. At step $i$, LU computes $\theta_i = U(\theta_{i-1},  F_1 \cup F_2 \dots \cup F_i, R \cup F_{i+1} \dots \cup F_k, \gamma_i):$ it unlearns $F_1 \cup F_2 \dots \cup F_i$ while retaining $R \cup F_{i+1} \dots \cup F_k$.

We analyze LU through inhibitors. In the $2$-fold case of unlearning tasks $A$ and $B$, the first stage forgets $A$ while retaining $B$, which forces the model to activate an inhibitor $I_A$ that selectively suppresses performance on $A$. In the second stage, the model may activate either an inhibitor $I_B$ that targets $B$ specifically, or a shared inhibitor $I_{AB}$ that suppresses both $A$ and $B$. 

Upon relearning $B$, any inhibitors affecting $B$—namely $I_B$ or $I_{AB}$—are deactivated, but $I_A$ remains active, so performance on $A$ stays suppressed. Conversely, when relearning $A$, the inhibitors $I_A$ and $I_{AB}$ are deactivated. If $I_{AB}$ had been activated, performance on $B$ is also restored; however, if $I_B$ had been activated instead, performance on $B$ remains suppressed. Whether the barrier to adversarial relearning is unidirectional or bidirectional depends on the unlearning primitive. $3$-fold LU is illustrated in Figure~\ref{fig:process}.

\begin{algorithm}[h]
\caption{Layered Unlearning}
\label{alg:seq_unlearning}
\begin{algorithmic}[1]
    \Require Model parameters $\theta_0$, forget dataset sequence $\{F_1, \dots, F_k\}$, retain dataset $R_0$, hyperparameters $\{\gamma_1, \dots, \gamma_k\}$, unlearning algorithm $U$
    \Ensure Unlearned model $\theta_k$

    \State $F = \emptyset, R = R_0 \bigcup_{i={1,\ldots,k}} F_i$ \Comment{Initialize incremental forget and retain sets $F$ and $R$.}
    
    \For{$i = 1$ to $k$} \Comment{Iterate through sequential forget stages}
        \State $F = F \cup F_i, R = R \setminus F_i$ \Comment{Update forget and retain sets}
        \State $\theta_i = U(\theta_{i-1}, F, R, \gamma_i)$ \Comment{Apply unlearning to this fold}
    \EndFor
    \State \Return $\theta_k$
\end{algorithmic}
\end{algorithm}

 %Each unlearning step proceeds until performance on that data fold drops below a forget threshold $T$. Appendix Section~\ref{appendix:hyperparameters} describes and discusses LU hyperparameters in detail. %For a graphical depiction of this process, see Appendix Figure~\ref{fig:rmu_wmdp_B_unlearn}. To determine when to proceed to the next stages, we use a forget threshold based on task accuracy; see Appendix Section~\ref{appendix:hyperparameters} for details.

%Our approach is broadly applicable to any unlearning method that defines both a forget set and a retain set, encompassing most existing unlearning techniques. For clarity, we denote \textit{Algorithm}-LU as an unlearning algorithm augmented with Layered Unlearning. For instance, RMU with Layered Unlearning is referred to as L-RMU.
\pagebreak
\subsubsection{Layered Unlearning for logistic regression}

\begin{wraptable}{r}{0.5\textwidth}
  \vspace{-1.3em}           % tune vertical placement
  \centering
  \caption{LU performance for logistic regression with random Gaussian assignment and $5$ Gaussians per dataset. ($10$ random seeds average).}
  \label{tab:gmm_acc}
  \begin{tabular}{@{} llccc @{}}
    \toprule
      Method & Relearn & A $\downarrow$ & B  $\downarrow$  & R $\uparrow$ \\
    \midrule
      Original & --- & 1.00 & 1.00 & 0.88 \\
      U        & --- &  0.02 & 0.01&0.96           \\
      U-LU       & --- & 0.01 & 0.01 & 0.96 \\
    \midrule
      U        & A   & --- & \bfseries 0.93 &   \bfseries 0.80 \\
      U-LU       & A   & --- & 0.96 & 0.78 \\
    \midrule
      U        & B   & 0.93 & --- &  0.80 \\
      U-LU       & B   & \bfseries 0.30 & --- & \bfseries 0.86 \\
    \bottomrule
  \end{tabular}
  \vspace{-3em}           % tune spacing below
\end{wraptable}

The bottom row of Figure~\ref{fig:gmm_heatmap} shows the sequence of weights that are generated by 2-fold LU. To illustrate the effect, we arrange $A, B, R$ as concentric circles with $A$ in the center. We can see that the most central weights $(\pm5,\pm5)$ decrease to forget $A$ while the surrounding circle of weights \emph{increases} to retain $B$. This distinction is preserved when $A\cup B$ is unlearned and so relearning $B$ does not recover performance on $A.$ With 5 Gaussians in the dataset, relearning on $B$ increases accuracy on $A$ by only $0.29$, compared to a $0.91$ increase in accuracy on $B$ when relearning on $A$ (see Table~\ref{tab:gmm_acc}).

We conducted experiments varying both the number of Gaussians in the mixture and the procedures used to assign the Gaussians to $A, B, R$. We find that increased task overlap leads to more adversarial relearning. This can occur by increasing the number of Gaussians per cluster (expanding the region of potential overlap) or by randomly assigning Gaussians to $A$, $B$, and $R$. To reduce overlap, we also cluster Gaussian means and assign entire clusters to $A$, $B$, and $R$, which significantly reduces adversarial relearning across all unlearning algorithms.

While the setup in Figure~\ref{fig:gmm_heatmap} is deliberately simplified to visualize inhibitors, we observe similar trends across all configurations (Appendix~\ref{appendix:synthetic}). In particular, when Gaussian means are randomly sampled—causing more overlap between $A$, $B$, and the retain task—standard unlearning becomes notably less robust. In contrast, when components are clustered to make folds more distinct, robustness improves, likely due to increased dataset separation making adversarial relearning more difficult.

\subsubsection{Layered Unlearning for bigram modeling}

% We implement the unlearning primitive $U$ by training the model to output a uniform distribution over tokens when conditioning on $F$.  We show the results of our experiment in Table~\ref{tab:bm_acc}. 
We measure the prediction performance for tasks $A, B$ with prediction accuracy. For the retain task $R$, we measure the total variation distance from a uniform distribution over $a, b$. We show the performance of the original weights, the unlearned weights with U and LU respectively, and the performance after relearning on $A$ or $B$. 

In this case, LU also confers bidirectional robustness to relearning generalization. While relearning $A$ or $B$ after U increases performance on the other task by $0.43$ on average. After LU, generalization accuracy only improves by $0.17$ on average. In contrast to our other experiments, we find that LU seems to activate fully independent inhibitors so that relearning does not transfer $A\rightarrow B$ or $B\rightarrow A$ (see Table~\ref{tab:bm_acc}).

% To better understand this robustness, we conduct ablations on transformer components (see Appendix~\ref{appendix:synthetic}). We find that components from U and LU models are largely interchangeable without affecting task performance, including the retain task. However, the attention components—specifically the $QK$ and $OV$ circuits—are critical for resisting adversarial relearning.

\begin{wraptable}{r}{0.5\textwidth}
  \vspace{-1.3em}           % tune vertical placement
  \centering
  \caption{LU performance for bigram language modeling with a 1-layer attention-only transformer. (Results averaged across 10 random seeds).}
  \label{tab:bm_acc}
  \begin{tabular}{@{} llccc @{}}
    \toprule
      Method & Relearn & A ↓    & B ↓    & R ↓ \\
    \midrule
      Original & --- & 0.91 & 0.91 & 0.02 \\
      U        & --- &  0.33 & 0.34 & 0.01           \\
      LU       & --- & 0.34 &  0.33 & 0.02 \\
    \midrule
      U        & A   & --- & 0.78 &  0.06 \\
      LU       & A   & --- & \bfseries 0.53 & \bfseries 0.04 \\
    \midrule
      U        & B   & 0.76 & --- &  0.05 \\
      LU       & B   & \bfseries 0.51 & --- & \bfseries 0.04 \\
    \bottomrule
  \end{tabular}
  \vspace{-1em}           % tune spacing below
\end{wraptable}

To better understand the source of LU's robustness, we conduct ablations on transformer components (see Appendix~\ref{appendix:synthetic}). We find that the components from U and LU models are interchangeable without affecting task performance, including on the retain set. However, the attention components—specifically the $QK$ and $OV$ circuits—are essential for resisting adversarial relearning, suggesting that robustness is encoded in the model’s attention mechanisms.

Consistent with this, retain set performance remains stable under relearning, indicating that the transformer does not revert to uniform predictions but instead applies targeted inhibition to Tasks $A$ and $B$. In contrast, a zero-layer, embedding-only transformer exhibits little to no adversarial relearning, highlighting the role of depth and attention in shaping inhibitor behavior for this setting. While we do not fully explain this result, we release all code and data to facilitate future interpretability research.

\section{LLM unlearning experiments}

Next, we evaluate the performance of LU on LLM unlearning benchmarks. Specifically, we consider unlearning on the WMDP~\citep{li2024wmdp}, MMLU~\citep{hendrycks2021measuringmassivemultitasklanguage}, and Years~\citep{deeb2025unlearningmethodsremoveinformation} datasets. WMDP consists of dangerous knowledge framed as multiple-choice questions (MCQs). To assess LU’s ability to remove capability-related information, we also apply unlearning to subsets of MMLU directly. The Years dataset contains major world events annotated with the year in which they occurred. For retain set evaluation, we use MMLU; when unlearning on MMLU, we exclude the categories being unlearned from the retain set.

% Next, we explore the performance of LU on LLM unlearning benchmarks. We consider unlearning on the WMDP~\citep{li2024wmdp}, MMLU \citep{hendrycks2021measuringmassivemultitasklanguage}, and Years~\citep{deeb2025unlearningmethodsremoveinformation} datasets. WMDP is a dataset of dangerous knowledge represented as multiple-choice questions (MCQ). To investigate the ability of LU to remove capability information, we also consider unlearning subsets of MMLU directly. Years is a dataset of major world events labeled with the year they occurred. For the retain set evaluation, we use MMLU~\citep{hendrycks2021measuringmassivemultitasklanguage}. 

\paragraph{Unlearning.} State-of-the-art LLM unlearning methods fall into two categories: representation engineering and gradient ascent. We select a representative algorithm from each for our unlearning primitive $U$: \emph{Representation Misdirection Unlearning} (RMU)~\citep{li2024wmdpbenchmarkmeasuringreducing} for representation engineering, and \emph{Simple Negative Policy Optimization} (SimNPO)~\citep{fan2025simplicityprevailsrethinkingnegative} for gradient ascent. We evaluate both methods with and without LU, denoting the LU variants as L-RMU and L-SimNPO, respectively. For a graphical overview of the unlearning process, see Appendix Figure~\ref{fig:rmu_wmdp_B_unlearn}.

\paragraph{Evaluation.}  
Given a dataset $F$ to be unlearned, we uniformly at random split it into $k$ folds, $F_1, \dots, F_k$. For a fixed $k$ and dataset, this partitioning remains consistent across all experiments for both unlearning and evaluation. We follow the Language Model Evaluation
Harness standards for $0$-shot evaluation \citep{eval-harness}.

To evaluate a model $\mathcal{M}$, we consider all $2^k - 2$ proper subsets $S \subset \{F_1, \dots, F_k\}$ and follow the evaluation protocol in \citet{deeb2025unlearningmethodsremoveinformation}. Specifically, we fine-tune $\mathcal{M}$ on either the MCQ data or the corresponding corpus from $S$ and then evaluate it on the MCQ questions from $T \coloneq F \setminus S$. We track accuracy on T over epochs and report the best accuracy as the final forget accuracy. All fine-tuning experiments use the Adam optimizer \citep{kingma2017adammethodstochasticoptimization}.

% Specifically, we analyze two RMU variants: RMU-LAT and RMU-Split. In RMU-Split, each fold is projected onto a different random vector, in contrast to the shared projection used in RMU. This isolates the impact of Layered Unlearning from the confounding effect of using separate random vectors.

% Our primary evaluation metric is the \textbf{recovery rate}. Consider two resulting models, $U$ and $V$, after unlearning. We aim to compare the percentage of information recovered by adversarial relearning on each model. We define the recovery rate as:
% \begin{equation*}
% \text{Recovery Rate} = \frac{\text{U Accuracy After Relearning} - \text{U Accuracy After Unlearning}}{\text{V Accuracy After Relearning} - \text{V Accuracy After Unlearning}}.
% \end{equation*}
% This metric quantifies how much more vulnerable model $U$ is to relearning compared to model $V$, with a lower recovery rate indicating better unlearning robustness.

% To ensure fair comparisons, we impose a minimum unlearning accuracy threshold of $0.25$ (random guessing for MCQs) in cases where the model is highly unlearned. This prevents artificially inflating the recovery rate when the model has already reached the theoretical lower bound of performance.

To ensure model utility, we only consider unlearned models that experience at most a $10\%$ accuracy drop on the retain set, see Appendix Table~\ref{tab:retain_acc}. All experiments are conducted using \textit{Zephyr-7B-$\beta$} \citep{tunstall2023zephyrdirectdistillationlm}.

\subsection{Layered Unlearning is more robust to adversarial relearning}

\begin{figure}[h]
\begin{center}
\includesvg[width=\textwidth]{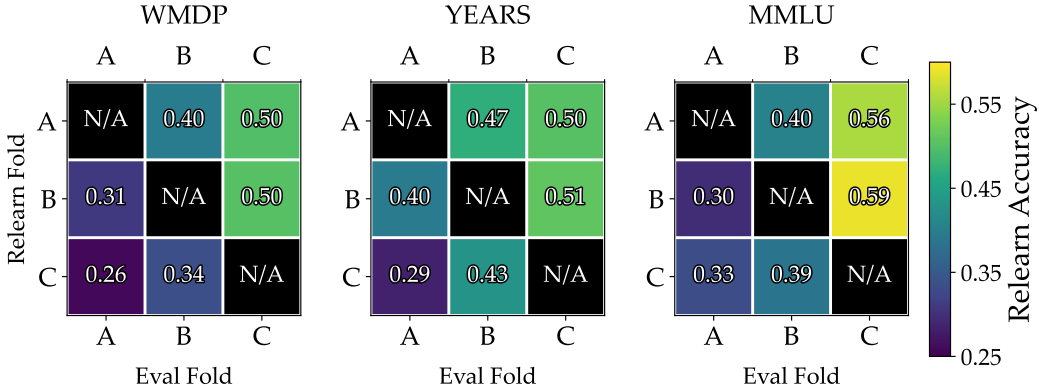}
\end{center}
\caption{
Model accuracy after relearning different folds of the data for an experiment with Layered RMU (L-RMU) and the folds $\{A, B, C\}$ in order. Each row shows the performance per fold for different relearning subsets. Notice that values below the diagonal are lower than values above the diagonal. This shows that L-RMU introduces a one-way barrier to relearning: relearning on $B$ regains performance on $C$ but not on $A$. 
}
\label{fig:order_heatmap}
%Relearn accuracy heatmaps across three datasets (WMDP, Years, MMLU). Each cell represents the relearning accuracy when training on one fold (y-axis) and evaluating on another (x-axis). We have relearning on MCQ in the first row and relearning on corpus in the second row. In all cases, values below the main diagonal are lower than values above the main diagonal, demonstrating the one-way barrier to relearning.
% Relearn accuracy heatmaps for L-RMU-Split across three datasets (WMDP, Years, MMLU). Each cell represents the relearning accuracy when training on one fold (y-axis) and evaluating on another (x-axis). Values range from $0.25$ to $0.75$ , with diagonal cells (black) representing invalid combinations where evaluated and relearned folds are identical. Higher values indicate more of the unlearned knowledge has been recovered. We have relearning on MCQ in the first row and relearning on corpus in the second row. 

\end{figure}

Across our experiments, we find that RMU becomes more robust to adversarial relearning when augmented with LU (Figure~\ref{fig:order_heatmap}). This robustness is sensitive to the order of folds: an adversary with access to fold $A$ can recover more information about $B$ and $C$ than one with access only to fold $C$. This reflects a path-dependent property of LU, where the sequence of unlearning influences the model's vulnerability to relearning.

SimNPO exhibits a similar, though weaker, improvement in robustness under LU. Notably, L-SimNPO produces a more symmetric barrier effect, in contrast to the directional robustness seen with RMU. These results indicate that LU generalizes across different unlearning methods.

% Consider the case where we unlearn two sets, $A$ and $B$, with $k = 2$. If unlearning were a path-independent operation, then sequentially unlearning $A$ while retaining $B$, followed by unlearning $B$, should yield the same result as unlearning $A$ and $B$ simultaneously. We exploit path-dependence to strategically design the order of forgetting to enhance robustness against adversarial relearning, leveraging the fact that different sequences of unlearning can lead to different retention and generalization behaviors in the final model.

% Compared to other tailored defenses to adversarial relearning, such as latent adversarial training (LAT)  \citep{sheshadri2025latent}, we find LU is consistently more robust for RMU (Table~\ref{tab:corpus_mcq_table}) when the relearned fold is unlearned after the evaluated fold. Otherwise, it performs comparably.  

\subsection{Corpus-based fine-tuning is a stronger adversarial attack}
We investigate the limits of LU by replacing MCQ-based prompts with corpus-based fine-tuning. While MCQ-based fine-tuning is commonly used due to its guaranteed performance improvement on targeted questions, corpus-based fine-tuning may more directly realign internal representations, making it a potentially stronger attack—particularly against unlearning methods based on representation engineering.

This substitution reveals a new state-of-the-art attack—one that only becomes apparent because LU enhances robustness to standard MCQ-based attacks. Although some robustness remains, it is reduced, as shown in Figure~\ref{fig:order_heatmap} and Table~\ref{tab:corpus_mcq_table}. Concretely, RMU's performance increases by an average of $5\%$ under corpus-based fine-tuning. For L-RMU, the effect is larger: performance improves by $10\%$ on average when relearning on later folds and evaluating on earlier ones, and by $6\%$ when relearning on earlier folds and evaluating on later ones. This asymmetry emerges only because LU introduces additional robustness, revealing corpus-based fine-tuning as a more effective attack in certain cases.

Interestingly, both SimNPO and L-SimNPO are less affected by corpus-based fine-tuning. However, SimNPO remains more vulnerable to adversarial relearning overall. This contrast shows that unlearning methods differ not only in their effectiveness, but also in the nature of their vulnerabilities. These findings highlight the importance of developing unlearning techniques that can withstand a diverse range of relearning attacks. 

% We speculate that corpus-based fine-tuning is particularly effective against unlearning methods that better preserve the model’s general language modeling capabilities, though this remains an open question for future research.

\begin{table}[ht]
\begin{center}
\begin{tabular}{clcccccc}
\toprule
\multirow{2}{*}{\bf Relearn} & \multirow{2}{*}{\bf Method} &\multicolumn{2}{c}{\bf A $\downarrow$} & \multicolumn{2}{c}{\bf B $\downarrow$} & \multicolumn{2}{c}{\bf C $\downarrow$} \\
\cmidrule(lr){3-4} \cmidrule(lr){5-6} \cmidrule(lr){7-8} 
& & MCQ & Corpus & MCQ & Corpus & MCQ & Corpus \\
\midrule
A & RMU  & --- & --- & 0.41 & 0.45 & 0.45 & 0.49 \\
A & L-RMU  & --- & --- & \bfseries 0.40 & 0.49 & 0.50 & 0.54 \\
A & SimNPO  & --- & --- & 0.47 & 0.45 & 0.50 & 0.54 \\
A & L-SimNPO  & --- & --- & 0.41 & \bf{0.35} & \bf{0.41} & \bf{0.40} \\
% \hdashline
% \noalign{\vskip 1mm} \multicolumn{2}{c}{Avg (Corpus - MCQ)} & \multicolumn{2}{c}{---} & \multicolumn{2}{c}{0.01} & \multicolumn{2}{c}{0.03} \\
\midrule
B & RMU  & 0.41 & 0.48 & --- & --- & 0.46 & 0.48 \\
B & L-RMU  & \bf{0.31} & \bf{0.44} & --- & --- & 0.50 & 0.54 \\
B & SimNPO  & 0.54 & 0.52 & --- & --- & 0.54 & 0.53 \\
B & L-SimNPO  & 0.42 & 0.45 & --- & --- & \bf{0.40} & \bf{0.41} \\
% \hdashline
% \noalign{\vskip 1mm}  \multicolumn{2}{c}{Avg (Corpus - MCQ)} & \multicolumn{2}{c}{0.05} & \multicolumn{2}{c}{---} &  \multicolumn{2}{c}{0.02} \\
\midrule
C & RMU  & 0.43 & 0.50 & 0.39 & 0.44 & --- & --- \\
C & L-RMU  & \bf{0.26} & \bf{0.36} & \bf{0.34} & \bf{0.42} & --- & --- \\
C & SimNPO  & 0.52 & 0.55 & 0.48 & 0.44 & --- & --- \\
C & L-SimNPO  & 0.42 & 0.47 & 0.45 & \bf{0.42} & --- & --- \\
% \hdashline
% \noalign{\vskip 1mm} \multicolumn{2}{c}{Avg (Corpus - MCQ)}  & \multicolumn{2}{c}{0.06} & \multicolumn{2}{c}{0.02} & \multicolumn{2}{c}{---}\\
\bottomrule
\end{tabular}
\end{center}
\caption{Relearning accuracies on WMDP for RMU, SimNPO, and 3-fold layered variants of both. We see that layered variants are more robust to relearning. This robustness is one-directional for L-RMU and partially bidirectional for L-SimNPO. This also shows that corpus attacks are generally more performance than multiple choice (MCQ) for the RMU variants. Similar results for Years and MMLU are shown in Appendix~\ref{appendix:relearn_acc}.}
\label{tab:corpus_mcq_table}
\end{table}

\section{Related work}

We briefly review the relevant literature.

\paragraph{Unlearning for LLMs.} 

Machine unlearning for large language models (LLMs) has become an active area of research~\citep{lu2022quark,jang2022knowledge,kumar2022privacy,zhang2023forget,pawelczyk2023context,eldan2023whos,ishibashi2023knowledge,yao2023large,maini2024tofu,zhang2024negative,li2024wmdp,wang2024large,jia2024soul,liu2024rethinking,liu2024large,thaker2024guardrail,kadhe2024split,fan2025simplicityprevailsrethinkingnegative,zhang2024unforgettablegeneralizationlanguagemodels}. Due to the difficulty of exact unlearning, most existing methods adopt approximate strategies, including model optimization~\citep{ilharco2022editing,liu2022continual,yao2023large,eldan2023whos,jia2024soul,zhang2024negative,li2024wmdp} and prompt-based or in-context learning techniques~\citep{thaker2024guardrail,pawelczyk2023context,liu2024large}. However, recent work has shown that these models often remain vulnerable to adversarial attacks~\citep{schwarzschild2024rethinking,patil2023can,lynch2024eight} or to relearning from small fragments of previously seen data~\citep{hu2024jogging,lynch2024eight}. These findings highlight the persistent challenges in achieving robust unlearning in LLMs.

\paragraph{Adversarial relearning.}
Adversarial relearning attacks exploit residual knowledge after unlearning by fine-tuning on a small subset of forgotten data, aiming to recover information about the full unlearned set. \citet{che2024model} showed that most existing unlearning methods are vulnerable to such attacks, revealing a fundamental limitation. \citet{deeb2025unlearningmethodsremoveinformation} further demonstrated that even informationally distinct examples can induce relearning, indicating failures beyond rote memorization. While several defenses have been proposed~\citep{rosati2024representationnoisingdefencemechanism, zou2024improvingalignmentrobustnesscircuit, tamirisa2025tamperresistant,sheshadri2025latent}, none have consistently withstood adversarial relearning~\citep{che2024model}. 
% Our method, like latent adversarial training (LAT)~\citep{sheshadri2025latent}, augments existing unlearning algorithms to improve robustness. 

Prior work has studied both corpus-based~\citep{che2024model} and MCQ-based~\citep{deeb2025unlearningmethodsremoveinformation} fine-tuning. To our knowledge, no comprehensive comparison of the two strategies has been conducted; we find corpus-based fine-tuning to be a more natural and effective form of adversarial relearning.

\paragraph{Sequential unlearning.}  
Sequential unlearning has been explored in various contexts, such as removing copyrighted information over time \citep{dou2025avoidingcopyrightinfringementlarge}. \citet{zhao2024makesunlearninghard} studied sequential unlearning as a means to improve forgetting efficiency but did not consider its impact on robustness against adversarial relearning. In contrast, our work investigates how the order and structure of sequential unlearning influence robustness, focusing on its potential to mitigate adversarial relearning. Specifically, we analyze the path dependence of unlearning and propose a novel framework that leverages structured forgetting to enhance resilience against information leakage.

\section{Discussion}

LU is sensitive to hyperparameters, requiring careful tuning at each stage to balance forgetting and retention. This sensitivity reflects a fundamental challenge in disentangling representations of forgotten and retained data. As the number of folds increases, the model must effectively discriminate between each pair of folds, with complexity scaling as $\binom{k}{2}$, which limits scalability to large $k$. Furthermore, LU is by nature more computationally intensive. However, in exchange for taking more time, it discovers optima that standard unlearning techniques are unable to discover no matter how long they train. Future work could consider more efficient methods.

Comprehensive adversarial evaluation is also difficult due to the exponential number of relearning subsets ($2^k - 2$) and additional attack configurations (e.g., batch size, learning rate, dataset, unlearning method). While we leave a full analysis to future work, our fixed hyperparameter setting was sufficient to break all baseline methods (RMU, SimNPO), as detailed in the Appendix.

We do not directly address the challenge of harmless fine-tuning, where the attacker uses data unrelated to what was unlearned, but we offer an intuition to guide future work. In standard unlearning, relearning is significantly easier when the attacker has access to data that is similar to the unlearned examples. Even small amounts of related data can serve as powerful signals, making it surprisingly effective to recover forgotten information. We hypothesize that LU reduces this vulnerability by making recovery difficult even when related data is available. As a result, it shrinks the performance gap between fine-tuning with related versus unrelated data, potentially making both equally ineffective.

Finally, we consider how the structure of LU might inform post-training more broadly. \citet{betley2025emergentmisalignmentnarrowfinetuning} show that fine-tuning on a single behavior—such as insecure code—can unintentionally induce harmful behaviors, suggesting entanglement between seemingly unrelated capabilities and values. One possible explanation is that a single post-training run introduces a shared inhibitor that influences multiple behaviors at once. LU, by contrast, creates multiple, distinct inhibitors and may help disentangle these behaviors. This perspective suggests a potential alignment strategy: first train a model to be harmless but helpless, then fine-tune it to be helpful while preserving harmlessness. In this setup, our results suggest increasing helplessness should preserve harmlessness, while increasing harmfulness should increase helplessness. While we focus on unlearning, we believe this layered approach could extend to alignment and other post-training interventions, offering a possible path toward more modular and controllable model behavior.

\section{Conclusion}
We introduced \textbf{Layered Unlearning}, a $k$-fold sequential unlearning framework that improves robustness by constructing functionally distinct, context-dependent inhibitors. By explicitly retaining a shrinking portion of the dataset at each stage, LU forces the model to localize forgetting, reducing the risk of shared failure under adversarial relearning. Our experiments demonstrate that LU reliably blocks recovery of earlier folds and significantly improves robustness across both synthetic and LLM benchmarks. While LU strengthens defenses against standard MCQ-based fine-tuning, it also reveals the limitations of current methods when faced with stronger corpus-based attacks. These results suggest that forgetting is inherently brittle and that robustness requires structured, layered defenses. More broadly, LU offers a testbed for mechanistic investigations of inhibitors and a conceptual foundation for more resilient post-training interventions.

\section{Acknowledgments}

This project's contributions were supported by Effective Giving and The Open Philanthropy Project.
% This work was supported by the National Science Foundation under grant IIS-2238240.

\bibliography{colm2025_conference}
\bibliographystyle{colm2025_conference}

\pagebreak

\appendix

\section{Layered Unlearning graphics}
We provide graphics to better communicate the main idea.

\begin{figure}[ht]
\begin{center}    
\includesvg[width=0.7\textwidth]{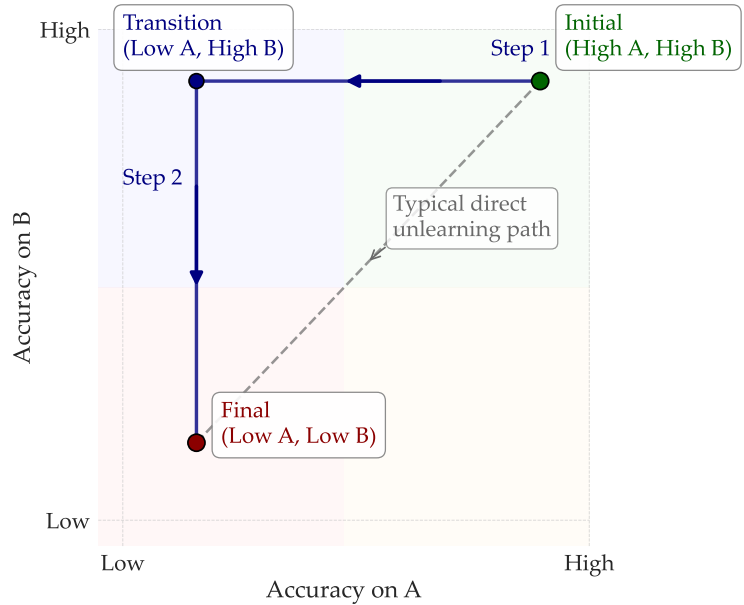}
\caption{The performance trajectory of LU on two folds $A, B$. Normally, unlearning methods lose performance on $A, B$ jointly and directly head towards the red point. However, we propose performing LU to retain performance on $B$ while forgetting $A$ and then forgetting both folds.}
\label{fig:unlearning_trajectory}
\end{center}
\end{figure}
\begin{figure}[ht]
\begin{center}
\includesvg[width=\textwidth]{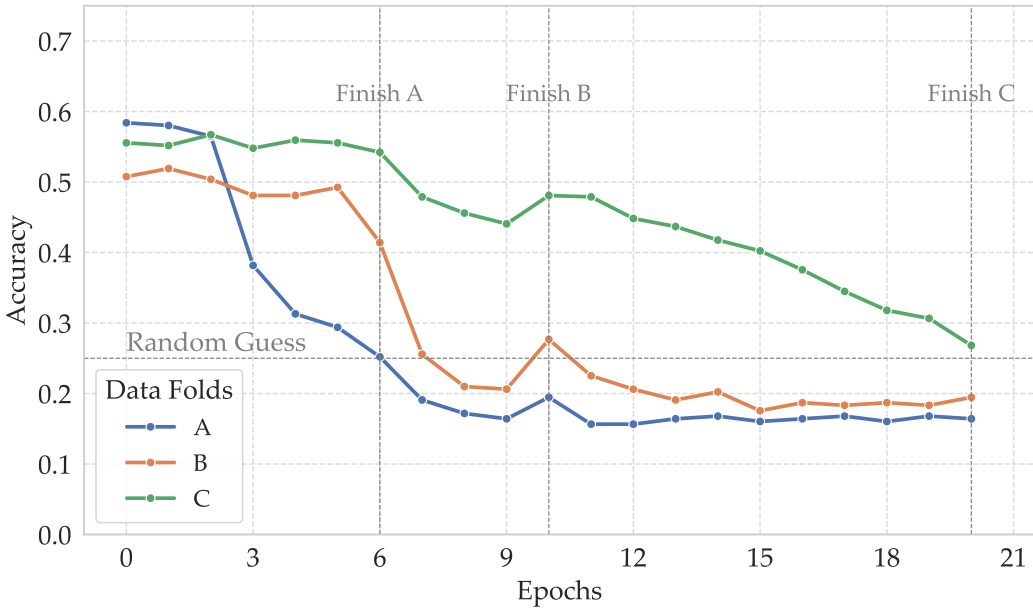}
\end{center}
\caption{We show the accuracy progression of forgetting three sets $A, B, C$ in that order using RMU on WMDP. The vertical dotted gray lines show when we move to forgetting the next fold. Note that the $A$ accuracy drops in the first iteration of forgetting and remains low. The accuracy of $B$ remains high until the second iteration of forgetting, and then drops and remains low. Finally, the accuracy of $C$ remains high until the third half of forgetting, when it drops.}
\label{fig:rmu_wmdp_B_unlearn}
\end{figure}

\pagebreak

\section{Synthetic ablation experiments}
\label{appendix:synthetic}

\subsection{Logistic regression}
\begin{figure}[ht]
    \centering
    \includesvg[width=0.8\textwidth]{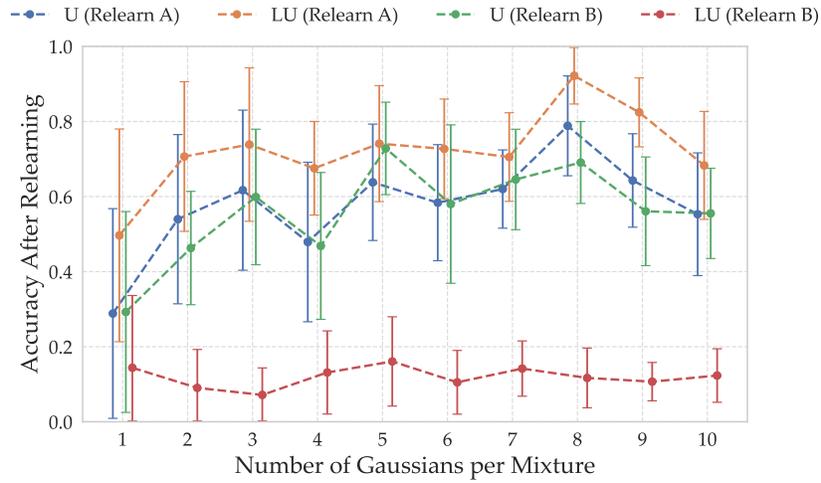}
    \caption{Relearning accuracies on $A$ and $B$ using datasets generated via K-Means clustering. Error bars denote $2$-std confidence intervals across $10$ random seeds. }
    \label{fig:gmm_k_means_AB}
\end{figure}

\begin{figure}[ht]
    \centering
    \includesvg[width=0.8\textwidth]{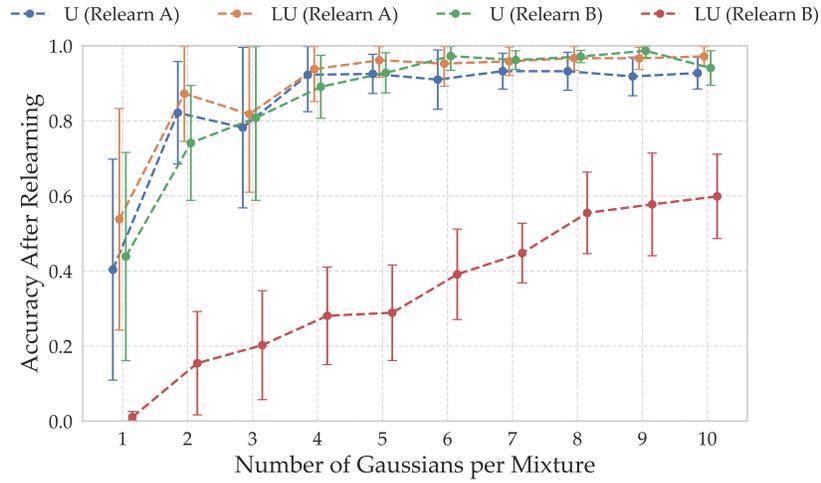}
    \caption{Relearning accuracies on $A$ and $B$ using datasets generated via random clustering. Error bars denote $2$-std confidence intervals across $10$ random seeds.}
    \label{fig:gmm_random_AB}
\end{figure}

We investigate different clustering schemes for grouping Gaussians into tasks $A$, $B$, and $R$. In the K-Means setup, we first cluster the Gaussian means using K-Means, then solve a linear assignment problem to evenly assign clusters to tasks based on proximity. Appendix Figures~\ref{fig:gmm_k_means_AB} and~\ref{fig:gmm_random_AB} show that adversarial relearning becomes more effective as the number of clusters increases. In contrast, LU consistently resists relearning, though its robustness is somewhat reduced under random clustering.

Our intuition is that adversarial relearning is less effective when task boundaries are more distinct. K-Means clustering tends to separate tasks more cleanly, thereby limiting overlap. In contrast, random clustering—especially with a larger number of Gaussians—increases the likelihood of overlap between tasks, making it more difficult to defend against adversarial relearning. Additionally, increasing the number of clusters inherently raises the potential for such overlap.


\subsection{Bigram modeling}

We analyze the effect of substituting components from the LU model into the U model on adversarial relearning, using the notation of \citet{elhage2021mathematical}. Substitutions are grouped as follows:
\begin{itemize}
    \item $QK$: Replace $W_Q$ and $W_K$.
    \item $OV$: Replace $W_O$ and $W_V$.
    \item $UE$: Replace $W_U$ and $W_E$.
\end{itemize}
These groupings reflect functional units in the model. As shown in Appendix Table~\ref{tab:synthetic_relearn}, substituting $QK$ or $OV$ consistently yields the greatest robustness to adversarial relearning, highlighting the key role of attention for this setting. This pattern is also visible in Appendix Figure~\ref{fig:bigram_accuracies}. Notably, retain and task accuracies remain stable across all substitution settings (Appendix Table~\ref{tab:synthetic_unlearn}), indicating no degradation in core performance. Investigating how attention confers this robustness is left for future work.

\begin{figure}[h]
    \centering
    \includesvg[width=0.8\textwidth]{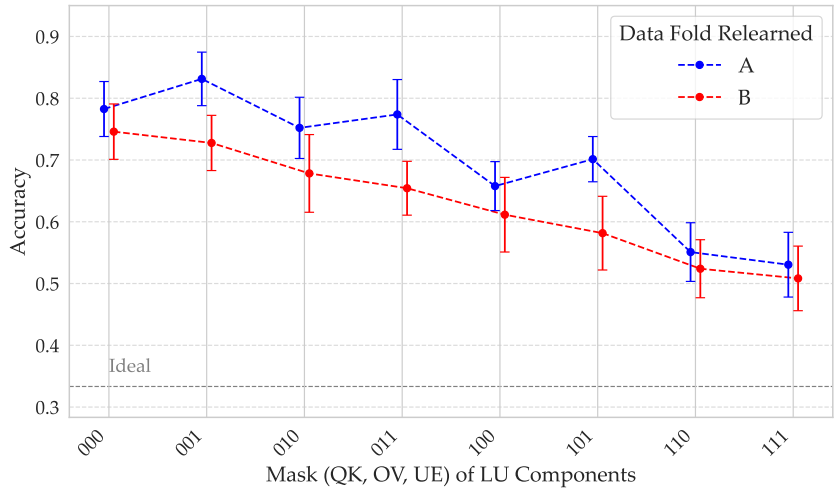}
    \caption{Relearning accuracies on $A$ and $B$ as a function of which components ($QK$, $OV$, $UE$) of the LU model are substituted in. The $x$-axis encodes binary masks (e.g., $000$ is standard unlearning; $111$ is full LU). Error bars denote $2$-std confidence intervals over $10$ random seeds. One seed was excluded due to universal resistance to relearning, which only increased error bar size without affecting trends. ``Ideal'' denotes perfect unlearning.}
    \label{fig:bigram_accuracies}
\end{figure}

\begin{table}[H]
\begin{center}
\begin{tabular}{ccccccc}
\toprule
\multicolumn{1}{c}{\bf Relearn} & \multicolumn{1}{c}{\bf QK} & \multicolumn{1}{c}{\bf OV} & \multicolumn{1}{c}{\bf UE} & \multicolumn{1}{c}{\bf A $\downarrow$} & \multicolumn{1}{c}{\bf B $\downarrow$} &  \multicolumn{1}{c}{\bf Retain $\downarrow$}\\ 
\midrule
A & 0 & 0 & 0 & --- & 0.78 & 0.06 \\
A & 0 & 0 & 1 & --- & 0.83 & 0.05 \\
A & 0 & 1 & 0 & --- & 0.75 & 0.05 \\
A & 0 & 1 & 1 & --- & 0.77 & 0.05 \\
A & 1 & 0 & 0 & --- & 0.66 & 0.04 \\
A & 1 & 0 & 1 & --- & 0.70 & \bf{0.03} \\
A & 1 & 1 & 0 & --- & 0.55 & \bf{0.03} \\
A & 1 & 1 & 1 & --- & \bf{0.53} & 0.04 \\
\midrule
B & 0 & 0 & 0 & 0.75 & --- & 0.05 \\
B & 0 & 0 & 1 & 0.73 & --- & 0.05 \\
B & 0 & 1 & 0 & 0.68 & --- & 0.05 \\
B & 0 & 1 & 1 & 0.65 & --- & \bf{0.04} \\
B & 1 & 0 & 0 & 0.61 & --- & 0.05 \\
B & 1 & 0 & 1 & 0.58 & --- & \bf{0.04} \\
B & 1 & 1 & 0 & 0.52 & --- & \bf{0.04} \\
B & 1 & 1 & 1 & \bf{0.51} & --- & \bf{0.04}   \\
\bottomrule
\end{tabular}
\end{center}
\caption{Table of relearning accuracies when substituting different model components. A value of $0$ indicates components from the U model, while $1$ indicates components from the LU model. The ``Retain'' column reports total variation (TV) distance on the retain set. This is an average over $10$ random seeds.}
\label{tab:synthetic_relearn}
\end{table}

\begin{table}[t]
\begin{center}
\begin{tabular}{cccccc}
\toprule
\multicolumn{1}{c}{\bf QK} & \multicolumn{1}{c}{\bf OV} & \multicolumn{1}{c}{\bf UE} & \multicolumn{1}{c}{\bf A $\downarrow$} & \multicolumn{1}{c}{\bf B $\downarrow$} &  \multicolumn{1}{c}{\bf Retain $\downarrow$}\\ 
\midrule
0 & 0 & 0 & 0.33 & 0.33 & 0.01\\
0 & 0 & 1 & 0.36 & 0.39 & 0.02 \\
0 & 1 & 0 & \bf{0.31} & \bf{0.30} & 0.02 \\
0 & 1 & 1 & 0.32 & 0.34 & 0.02 \\
1 & 0 & 0 & 0.33 & 0.33 & \bf{0.01} \\
1 & 0 & 1 & 0.36 & 0.39 & 0.02 \\
1 & 1 & 0 & 0.32 & \bf{0.30} & 0.02 \\
1 & 1 & 1 & 0.34 & 0.33 & \bf{0.02} \\
\bottomrule
\end{tabular}
\end{center}
\caption{Unlearned accuracies after substituting model components. A value of $0$ indicates components from the U model, and $1$ indicates components from the LU model. The ``Retain'' column reports the total variation (TV) distance on the retain set. This is an average over $10$ random seeds.}
\label{tab:synthetic_unlearn}
\end{table}
\pagebreak
\section{Retain accuracies}

For evaluation, we either use full MMLU evaluation or we use specific categories for MMLU created in \citet{deeb2025unlearningmethodsremoveinformation}. The retain categories for MMLU consist of questions relating to health, history, law, philosophy, and the social sciences. The forget categories for MMLU consist of questions relating to geography, culture, STEM, chemistry, and business. 

\begin{table}[H]
\begin{center}
\begin{tabular}{lclc}
\toprule
\multicolumn{1}{l}{\bf Size} & \multicolumn{1}{c}{\bf Dataset}  & \multicolumn{1}{l}{\bf Method} & \multicolumn{1}{c}{\bf Retain $\uparrow$}  \\ 
\midrule
Small & --- & None & 0.59 \\
Full & --- & None  & 0.58 \\
\midrule
Full & WMDP 2 & RMU  & 0.57 \\
Full & WMDP 2 & RMU-Split  & 0.58 \\
Full & WMDP 2 & L-RMU  & 0.57 \\
Full & WMDP 2 & L-RMU-Split  & 0.57 \\
\midrule
Full & WMDP 3 & RMU  & 0.57 \\
% Full & WMDP 3 & RMU-LAT  & 0.56 \\
Full & WMDP 3 & RMU-Split  & 0.57 \\
Full & WMDP 3 & L-RMU  & 0.57 \\
Full & WMDP 3 & L-RMU-Split  & 0.55 \\
Full & WMDP 3 & SimNPO  & 0.57 \\
Full & WMDP 3 & L-SimNPO  & 0.55 \\
\midrule
Full & WMDP 4 & RMU  & 0.57 \\
Full & WMDP 4 & RMU-Split  & 0.57 \\
Full & WMDP 4 & L-RMU  & 0.57 \\
Full & WMDP 4 & L-RMU-Split  & 0.57 \\
\midrule
Small & MMLU 3 & RMU  & 0.55 \\
Small & MMLU 3 & RMU-Split  & 0.56 \\
Small & MMLU 3 & L-RMU  & 0.54 \\
Small & MMLU 3 & L-RMU-Split  & 0.55 \\
\midrule
Full & Years 3 & RMU  & 0.58 \\
Full & Years 3 & RMU-Split  & 0.58 \\
Full & Years 3 & L-RMU  & 0.58 \\
Full & Years 3 & L-RMU-Split  & 0.57 \\
\bottomrule
\end{tabular}
\end{center}
\caption{Retain accuracy by method. We refer to the restricted subset of MMLU questions as the \textit{small set} and the full MMLU dataset as the \textit{full set}. The dataset column indicates which dataset was unlearned for each method. The first section shows results from the original model before any unlearning was applied.}
\label{tab:retain_acc}
\end{table}

\section{Unlearning accuracies}

We provide the accuracies after applying the unlearning methods on each fold for each dataset. We also analyze another variant of RMU, which we term RMU-Split. In RMU-Split, each fold is projected onto a different random vector, in contrast to the shared projection used in RMU. This isolates the impact of LU from the confounding effect of using separate random vectors. The results do not change.

\begin{table}[H]
\begin{center}
\begin{tabular}{llccccc}
\toprule
\multicolumn{1}{l}{\bf Dataset} & \multicolumn{1}{l}{\bf Method} & \multicolumn{1}{c}{\bf A $\downarrow$} & \multicolumn{1}{c}{\bf B $\downarrow$} & \multicolumn{1}{c}{\bf C $\downarrow$} & \multicolumn{1}{c}{\bf D $\downarrow$} \\ 
\midrule
WMDP 2 & RMU & 0.26 & 0.29 & --- & --- \\ 
WMDP 2 & L-RMU & \bf{0.21} & 0.28 & --- & --- \\ 
WMDP 2 & RMU-Split & 0.25 & \bf{0.26} & --- & --- \\ 
WMDP 2 & L-RMU-Split & 0.24 & 0.28 & --- & --- \\ 
\midrule
WMDP 3 & RMU & 0.27 & 0.24 & 0.33 & --- \\ 
% WMDP 3 & RMU-LAT & 0.21 & 0.19 & 0.25 & --- \\ 
WMDP 3 & L-RMU & \bf{0.15} & 0.24 & 0.33 & --- \\ 
WMDP 3 & RMU-Split & 0.22 & \bf{0.18} & \bf{0.22} & --- \\ 
WMDP 3 & L-RMU-Split & 0.16 & 0.19 & 0.27 & --- \\ 
WMDP 3 & SimNPO & 0.32 & 0.24 & 0.31 & --- \\ 
WMDP 3 & L-SimNPO & 0.29 & 0.32 & 0.33 & --- \\ 
\midrule
WMDP 4 & RMU & 0.25 & 0.27 & 0.29 & 0.30 \\ 
WMDP 4 & L-RMU & \bf{0.15} & \bf{0.21} & 0.25 & 0.34 \\ 
WMDP 4 & RMU-Split & 0.28 & 0.26 & \bf{0.23} & \bf{0.22} \\ 
WMDP 4 & L-RMU-Split & 0.18 & 0.23 & 0.27 & 0.34 \\ 
\midrule
MMLU 3 & RMU & 0.28 & 0.30 & \bf{0.30} & --- \\ 
MMLU 3 & L-RMU & 0.23 & 0.29 & 0.34 & --- \\ 
MMLU 3 & RMU-Split & 0.26 & 0.34 & 0.30 & --- \\ 
MMLU 3 & L-RMU-Split & \bf{0.21} & \bf{0.29} & 0.32 & --- \\ 
\midrule
Years 3 & RMU & 0.30 & 0.22 & 0.30 & --- \\ 
Years 3 & L-RMU & \bf{0.29} & 0.28 & 0.31 & --- \\ 
Years 3 & RMU-Split & 0.33 & 0.29 & \bf{0.22} & --- \\ 
Years 3 & L-RMU-Split & 0.29 & \bf{0.20} & 0.27 & --- \\ 
\bottomrule
\end{tabular}
\end{center}
\caption{Unlearning accuracy by method across datasets and evaluation folds. }
\label{tab:unlearn_acc}
\end{table}

\section{Relearning accuracies}
\label{appendix:relearn_acc}

All attacks are performed with learning rate $10^{-6}$ and batch size $4$. We fine-tune on MCQ for $8$ epochs and fine-tune on corpus for $5$ epochs. This difference is because we wish to fine-tune until the accuracy on the relearn set stops increasing for a while, and by definition fine-tuning on MCQ can reach $1.0$ accuracy, so we fine-tune for longer. We then take the maximum validation accuracy across all epochs.

\subsection{WMDP 2 folds}
\begin{table}[H]
\caption{Relearning accuracy across methods for WMDP 2 folds.}
\label{tab:wmdp_2_acc_relearn}
\begin{center}
\begin{tabular}{clcccc}
\toprule
\multirow{2}{*}{\bf Relearn} & \multirow{2}{*}{\bf Method} &\multicolumn{2}{c}{\bf A $\downarrow$} & \multicolumn{2}{c}{\bf B $\downarrow$} \\
\cmidrule(lr){3-4} \cmidrule(lr){5-6} 
 & & MCQ & Corpus & MCQ & Corpus \\
\midrule
 A & RMU  & --- & --- & 0.45 & \bf{0.48} \\
 A & L-RMU  & --- & --- & 0.43 & 0.53 \\
 A & RMU-Split  & --- & --- & \bf{0.42} & 0.51 \\
     A & L-RMU-Split  & --- & --- & 0.42 & 0.54 \\
\midrule
 B & RMU  & 0.45 & 0.51 & --- & --- \\
 B & L-RMU  & \bf{0.30} & \bf{0.41} & --- & --- \\
 B & RMU-Split  & 0.40 & 0.53 & --- & --- \\
 B & L-RMU-Split  & 0.32 & 0.48 & --- & --- \\
\bottomrule
\end{tabular}
\end{center}
\end{table}

\pagebreak
\subsection{WMDP 3 folds}
\begin{table}[H]
\caption{Relearning accuracy across methods for WMDP 3 folds.}
\label{tab:wmdp_3_acc_relearn}
\begin{center}
\begin{tabular}{clcccccc}
\toprule
\multirow{2}{*}{\bf Relearn} & \multirow{2}{*}{\bf Method} &\multicolumn{2}{c}{\bf A $\downarrow$} & \multicolumn{2}{c}{\bf B $\downarrow$} & \multicolumn{2}{c}{\bf C $\downarrow$} \\
\cmidrule(lr){3-4} \cmidrule(lr){5-6} \cmidrule(lr){7-8} 
 & & MCQ & Corpus & MCQ & Corpus & MCQ & Corpus \\
\midrule
 A & RMU  & --- & --- & 0.41 & 0.45 & 0.45 & 0.49 \\
 % A & RMU-LAT  & --- & --- & 0.39 & 0.44 & 0.48 & 0.55 \\
 A & L-RMU  & --- & --- & 0.40 & 0.49 & 0.50 & 0.54 \\
 A & RMU-Split  & --- & --- & 0.37 & 0.44 & \bf{0.38} & 0.54 \\
 A & L-RMU-Split  & --- & --- & \bf{0.35} & 0.45 & 0.49 & 0.55 \\
 A & SimNPO  & --- & --- & 0.47 & 0.45 & 0.50 & 0.54 \\
 A & L-SimNPO  & --- & --- & 0.41 & \bf{0.35} & 0.41 & \bf{0.40} \\
\midrule
 B & RMU  & 0.41 & 0.48 & --- & --- & 0.46 & 0.48 \\
 % B & RMU-LAT  & 0.40 & 0.48 & --- & --- & 0.44 & 0.48 \\
 B & L-RMU  & 0.31 & 0.44 & --- & --- & 0.50 & 0.54 \\
 B & RMU-Split  & 0.42 & 0.53 & --- & --- & \bf{0.38} & 0.54 \\
 B & L-RMU-Split  & \bf{0.29} & \bf{0.43} & --- & --- & 0.47 & 0.54 \\
 B & SimNPO  & 0.54 & 0.52 & --- & --- & 0.54 & 0.53 \\
 B & L-SimNPO  & 0.42 & 0.45 & --- & --- & 0.40 & \bf{0.41} \\
\midrule
 C & RMU  & 0.43 & 0.50 & 0.39 & 0.44 & --- & --- \\
 % C & RMU-LAT  & 0.39 & 0.50 & 0.42 & 0.43 & --- & --- \\
 C & L-RMU  & 0.26 & 0.36 & 0.34 & 0.42 & --- & --- \\
 C & RMU-Split  & 0.40 & 0.55 & 0.39 & 0.45 & --- & --- \\
 C & L-RMU-Split  & \bf{0.25} & \bf{0.29} & \bf{0.27} & \bf{0.34} & --- & --- \\
 C & SimNPO  & 0.52 & 0.55 & 0.48 & 0.44 & --- & --- \\
 C & L-SimNPO  & 0.42 & 0.47 & 0.45 & 0.42 & --- & --- \\
\midrule
 A, B & RMU  & --- & --- & --- & --- & 0.48 & 0.51 \\
 % A, B & RMU-LAT  & --- & --- & --- & --- & 0.51 & 0.56 \\
 A, B & L-RMU  & --- & --- & --- & --- & 0.49 & 0.56 \\
 A, B & RMU-Split  & --- & --- & --- & --- & \bf{0.41} & 0.55 \\
 A, B & L-RMU-Split  & --- & --- & --- & --- & 0.50 & 0.54 \\
 A, B & SimNPO  & --- & --- & --- & --- & 0.57 & 0.53 \\
 A, B & L-SimNPO  & --- & --- & --- & --- & 0.43 & \bf{0.40} \\
\midrule
 A, C & RMU  & --- & --- & 0.43 & 0.47 & --- & --- \\
 % A, C & RMU-LAT  & --- & --- & 0.40 & 0.47 & --- & --- \\
 A, C & L-RMU  & --- & --- & 0.38 & 0.50 & --- & --- \\
 A, C & RMU-Split  & --- & --- & 0.42 & 0.45 & --- & --- \\
 A, C & L-RMU-Split  & --- & --- & \bf{0.34} & 0.45 & --- & --- \\
 A, C & SimNPO  & --- & --- & 0.48 & 0.44 & --- & --- \\
 A, C & L-SimNPO  & --- & --- & 0.44 & \bf{0.40} & --- & --- \\
\midrule
 B, C & RMU  & 0.43 & 0.54 & --- & --- & --- & --- \\
 % B, C & RMU-LAT  & 0.39 & 0.55 & --- & --- & --- & --- \\
 B, C & L-RMU  & \bf{0.34} & 0.47 & --- & --- & --- & --- \\
 B, C & RMU-Split  & 0.39 & 0.56 & --- & --- & --- & --- \\
 B, C & L-RMU-Split  & 0.35 & \bf{0.42} & --- & --- & --- & --- \\
 B, C & SimNPO  & 0.52 & 0.56 & --- & --- & --- & --- \\
 B, C & L-SimNPO  & 0.44 & 0.47 & --- & --- & --- & --- \\
\bottomrule
\end{tabular}
\end{center}
\end{table}

\pagebreak
\subsection{WMDP 4 folds}
\begin{longtable}{clcccccccc}
\caption{Relearning accuracy across methods for WMDP 4 folds.}\\
\label{tab:wmdp_4_acc_relearn}\\
\toprule
\multirow{2}{*}{\bf Relearn} & \multirow{2}{*}{\bf Method} &\multicolumn{2}{c}{\bf A $\downarrow$} & \multicolumn{2}{c}{\bf B $\downarrow$} & \multicolumn{2}{c}{\bf C $\downarrow$} & \multicolumn{2}{c}{\bf D $\downarrow$} \\
\cmidrule(lr){3-4} \cmidrule(lr){5-6} \cmidrule(lr){7-8} \cmidrule(lr){9-10} 
 & & MCQ & Corpus & MCQ & Corpus & MCQ & Corpus & MCQ & Corpus \\
\midrule
\endfirsthead
\multicolumn{10}{c}{\textit{Continued from previous page}} \\
\\[1mm] % Adjust the space as needed
\toprule
\multirow{2}{*}{\bf Relearn} & \multirow{2}{*}{\bf Method} &\multicolumn{2}{c}{\bf A $\downarrow$} & \multicolumn{2}{c}{\bf B $\downarrow$} & \multicolumn{2}{c}{\bf C $\downarrow$} & \multicolumn{2}{c}{\bf D $\downarrow$} \\
\cmidrule(lr){3-4} \cmidrule(lr){5-6} \cmidrule(lr){7-8} \cmidrule(lr){9-10} 
 & & MCQ & Corpus & MCQ & Corpus & MCQ & Corpus & MCQ & Corpus \\
\midrule
\endhead
\\[1mm] % Adjust the space as needed
\multicolumn{10}{c}{\textit{Continued on next page}} \\
\endfoot
\bottomrule
\endlastfoot
 A & RMU  & --- & --- & 0.41 & \bf{0.44} & 0.37 & \bf{0.44} & 0.41 & \bf{0.47} \\
 A & L-RMU  & --- & --- & 0.38 & 0.46 & \bf{0.35} & 0.46 & \bf{0.40} & 0.57 \\
 A & RMU-Split  & --- & --- & 0.42 & 0.46 & 0.38 & 0.44 & 0.41 & 0.51 \\
 A & L-RMU-Split  & --- & --- & \bf{0.38} & 0.50 & 0.39 & 0.47 & 0.44 & 0.56 \\
\midrule
 B & RMU  & 0.46 & \bf{0.47} & --- & --- & 0.44 & \bf{0.41} & 0.49 & \bf{0.44} \\
 B & L-RMU  & \bf{0.29} & 0.51 & --- & --- & \bf{0.36} & 0.46 & 0.46 & 0.56 \\
 B & RMU-Split  & 0.42 & 0.51 & --- & --- & 0.37 & 0.45 & \bf{0.40} & 0.51 \\
 B & L-RMU-Split  & 0.33 & \bf{0.47} & --- & --- & 0.37 & 0.47 & 0.51 & 0.56 \\
\midrule
 C & RMU  & 0.44 & 0.49 & 0.43 & 0.47 & --- & --- & 0.46 & \bf{0.45} \\
 C & L-RMU  & 0.33 & \bf{0.45} & 0.37 & 0.47 & --- & --- & 0.46 & 0.57 \\
 C & RMU-Split  & 0.43 & 0.51 & \bf{0.35} & \bf{0.46} & --- & --- & \bf{0.41} & 0.49 \\
 C & L-RMU-Split  & \bf{0.29} & 0.47 & \bf{0.35} & 0.47 & --- & --- & 0.45 & 0.55 \\
\midrule
 D & RMU  & 0.43 & 0.47 & 0.42 & 0.47 & 0.40 & 0.42 & --- & --- \\
 D & L-RMU  & \bf{0.25} & \bf{0.33} & \bf{0.27} & \bf{0.42} & 0.36 & 0.47 & --- & --- \\
 D & RMU-Split  & 0.37 & 0.52 & 0.38 & 0.45 & 0.34 & 0.45 & --- & --- \\
 D & L-RMU-Split  & \bf{0.25} & 0.40 & 0.32 & 0.43 & \bf{0.32} & \bf{0.40} & --- & --- \\
\midrule
 A, B & RMU  & --- & --- & --- & --- & 0.41 & 0.47 & 0.48 & \bf{0.53} \\
 A, B & L-RMU  & --- & --- & --- & --- & 0.43 & 0.49 & 0.46 & 0.56 \\
 A, B & RMU-Split  & --- & --- & --- & --- & \bf{0.38} & \bf{0.45} & \bf{0.43} & \bf{0.53} \\
 A, B & L-RMU-Split  & --- & --- & --- & --- & 0.39 & 0.48 & 0.57 & 0.56 \\
\midrule
 A, C & RMU  & --- & --- & 0.41 & 0.49 & --- & --- & 0.50 & \bf{0.49} \\
 A, C & L-RMU  & --- & --- & 0.43 & 0.48 & --- & --- & 0.46 & 0.57 \\
 A, C & RMU-Split  & --- & --- & \bf{0.38} & \bf{0.48} & --- & --- & \bf{0.39} & 0.53 \\
 A, C & L-RMU-Split  & --- & --- & 0.41 & 0.52 & --- & --- & 0.48 & 0.56 \\
\midrule
 A, D & RMU  & --- & --- & 0.43 & 0.49 & 0.40 & \bf{0.45} & --- & --- \\
 A, D & L-RMU  & --- & --- & \bf{0.35} & \bf{0.48} & \bf{0.38} & 0.46 & --- & --- \\
 A, D & RMU-Split  & --- & --- & 0.46 & 0.49 & 0.41 & 0.47 & --- & --- \\
 A, D & L-RMU-Split  & --- & --- & 0.44 & 0.51 & 0.40 & 0.48 & --- & --- \\
\midrule
 B, C & RMU  & 0.47 & 0.53 & --- & --- & --- & --- & 0.48 & \bf{0.51} \\
 B, C & L-RMU  & \bf{0.30} & 0.53 & --- & --- & --- & --- & 0.46 & 0.56 \\
 B, C & RMU-Split  & 0.43 & 0.55 & --- & --- & --- & --- & \bf{0.40} & 0.53 \\
 B, C & L-RMU-Split  & 0.35 & \bf{0.49} & --- & --- & --- & --- & 0.56 & 0.56 \\
\midrule
 B, D & RMU  & 0.45 & 0.56 & --- & --- & 0.44 & \bf{0.46} & --- & --- \\
 B, D & L-RMU  & 0.35 & 0.51 & --- & --- & 0.42 & 0.47 & --- & --- \\
 B, D & RMU-Split  & 0.45 & 0.55 & --- & --- & \bf{0.39} & \bf{0.46} & --- & --- \\
 B, D & L-RMU-Split  & \bf{0.34} & \bf{0.49} & --- & --- & 0.39 & 0.49 & --- & --- \\
\midrule
 C, D & RMU  & 0.43 & 0.54 & 0.42 & 0.48 & --- & --- & --- & --- \\
 C, D & L-RMU  & \bf{0.27} & \bf{0.47} & \bf{0.35} & \bf{0.45} & --- & --- & --- & --- \\
 C, D & RMU-Split  & 0.36 & 0.55 & 0.38 & 0.48 & --- & --- & --- & --- \\
 C, D & L-RMU-Split  & 0.29 & 0.48 & 0.41 & 0.51 & --- & --- & --- & --- \\
\midrule
 A, B, C & RMU  & --- & --- & --- & --- & --- & --- & 0.46 & \bf{0.52} \\
 A, B, C & L-RMU  & --- & --- & --- & --- & --- & --- & 0.49 & 0.56 \\
 A, B, C & RMU-Split  & --- & --- & --- & --- & --- & --- & \bf{0.42} & 0.53 \\
 A, B, C & L-RMU-Split  & --- & --- & --- & --- & --- & --- & 0.59 & 0.56 \\
\midrule
 A, B, D & RMU  & --- & --- & --- & --- & 0.44 & 0.48 & --- & --- \\
 A, B, D & L-RMU  & --- & --- & --- & --- & \bf{0.42} & 0.51 & --- & --- \\
 A, B, D & RMU-Split  & --- & --- & --- & --- & \bf{0.42} & \bf{0.47} & --- & --- \\
 A, B, D & L-RMU-Split  & --- & --- & --- & --- & 0.45 & 0.49 & --- & --- \\
\midrule
 A, C, D & RMU  & --- & --- & 0.48 & 0.49 & --- & --- & --- & --- \\
 A, C, D & L-RMU  & --- & --- & \bf{0.39} & \bf{0.48} & --- & --- & --- & --- \\
 A, C, D & RMU-Split  & --- & --- & 0.49 & 0.49 & --- & --- & --- & --- \\
 A, C, D & L-RMU-Split  & --- & --- & 0.43 & 0.51 & --- & --- & --- & --- \\
\midrule
 B, C, D & RMU  & 0.47 & 0.55 & --- & --- & --- & --- & --- & --- \\
 B, C, D & L-RMU  & \bf{0.31} & 0.51 & --- & --- & --- & --- & --- & --- \\
 B, C, D & RMU-Split  & 0.44 & 0.57 & --- & --- & --- & --- & --- & --- \\
 B, C, D & L-RMU-Split  & 0.37 & \bf{0.48} & --- & --- & --- & --- & --- & --- \\
\end{longtable}

\subsection{MMLU 3 folds}
\begin{table}[H]
\caption{Relearning accuracy across methods for MMLU 3 folds.}
\label{tab:mmlu_3_acc_relearn}
\begin{center}
\begin{tabular}{clcccccc}
\toprule
\multirow{2}{*}{\bf Relearn} & \multirow{2}{*}{\bf Method} &\multicolumn{2}{c}{\bf A $\downarrow$} & \multicolumn{2}{c}{\bf B $\downarrow$} & \multicolumn{2}{c}{\bf C $\downarrow$} \\
\cmidrule(lr){3-4} \cmidrule(lr){5-6} \cmidrule(lr){7-8} 
 & & MCQ & Corpus & MCQ & Corpus & MCQ & Corpus \\
\midrule
 A & RMU  & --- & --- & 0.49 & 0.66 & 0.51 & 0.64 \\
 A & L-RMU  & --- & --- & \bf{0.40} & 0.61 & 0.56 & 0.64 \\
 A & RMU-Split  & --- & --- & 0.52 & 0.63 & \bf{0.48} & \bf{0.62} \\
 A & L-RMU-Split  & --- & --- & \bf{0.40} & \bf{0.57} & 0.50 & 0.64 \\
\midrule
 B & RMU  & 0.52 & 0.65 & --- & --- & 0.51 & \bf{0.62} \\
 B & L-RMU  & \bf{0.30} & \bf{0.54} & --- & --- & 0.59 & 0.64 \\
 B & RMU-Split  & 0.50 & 0.63 & --- & --- & \bf{0.49} & 0.62 \\
 B & L-RMU-Split  & 0.38 & 0.61 & --- & --- & \bf{0.49} & 0.63 \\
\midrule
 C & RMU  & 0.53 & 0.64 & 0.52 & 0.64 & --- & --- \\
 C & L-RMU  & \bf{0.33} & \bf{0.39} & 0.39 & \bf{0.46} & --- & --- \\
 C & RMU-Split  & 0.55 & 0.65 & 0.57 & 0.64 & --- & --- \\
 C & L-RMU-Split  & 0.36 & 0.48 & \bf{0.38} & 0.50 & --- & --- \\
\midrule
 A, B & RMU  & --- & --- & --- & --- & 0.54 & 0.63 \\
 A, B & L-RMU  & --- & --- & --- & --- & 0.60 & 0.64 \\
 A, B & RMU-Split  & --- & --- & --- & --- & 0.61 & 0.64 \\
 A, B & L-RMU-Split  & --- & --- & --- & --- & \bf{0.53} & \bf{0.63} \\
\midrule
 A, C & RMU  & --- & --- & 0.55 & 0.64 & --- & --- \\
 A, C & L-RMU  & --- & --- & \bf{0.44} & 0.61 & --- & --- \\
 A, C & RMU-Split  & --- & --- & 0.61 & 0.66 & --- & --- \\
 A, C & L-RMU-Split  & --- & --- & 0.48 & \bf{0.59} & --- & --- \\
\midrule
 B, C & RMU  & 0.52 & 0.65 & --- & --- & --- & --- \\
 B, C & L-RMU  & \bf{0.32} & \bf{0.55} & --- & --- & --- & --- \\
 B, C & RMU-Split  & 0.53 & 0.67 & --- & --- & --- & --- \\
 B, C & L-RMU-Split  & 0.38 & 0.60 & --- & --- & --- & --- \\
\bottomrule
\end{tabular}
\end{center}
\end{table}

\pagebreak
\subsection{Years 3 folds}
\begin{table}[H]
\caption{Relearning accuracy across methods for Years 3 folds.}
\label{tab:years_3_acc_relearn}
\begin{center}
\begin{tabular}{clcccccc}
\toprule
\multirow{2}{*}{\bf Relearn} & \multirow{2}{*}{\bf Method} &\multicolumn{2}{c}{\bf A $\downarrow$} & \multicolumn{2}{c}{\bf B $\downarrow$} & \multicolumn{2}{c}{\bf C $\downarrow$} \\
\cmidrule(lr){3-4} \cmidrule(lr){5-6} \cmidrule(lr){7-8} 
 & & MCQ & Corpus & MCQ & Corpus & MCQ & Corpus \\
\midrule
 A & RMU  & --- & --- & 0.55 & 0.57 & 0.55 & 0.58 \\
 A & L-RMU  & --- & --- & 0.47 & 0.45 & \bf{0.50} & 0.53 \\
 A & RMU-Split  & --- & --- & 0.55 & 0.48 & 0.54 & \bf{0.48} \\
 A & L-RMU-Split  & --- & --- & \bf{0.43} & \bf{0.36} & 0.51 & 0.50 \\
\midrule
 B & RMU  & 0.56 & 0.59 & --- & --- & 0.55 & 0.58 \\
 B & L-RMU  & 0.40 & 0.33 & --- & --- & 0.51 & \bf{0.51} \\
 B & RMU-Split  & 0.59 & 0.48 & --- & --- & \bf{0.49} & 0.52 \\
 B & L-RMU-Split  & \bf{0.37} & \bf{0.33} & --- & --- & 0.52 & 0.52 \\
\midrule
 C & RMU  & 0.54 & 0.58 & 0.58 & 0.58 & --- & --- \\
 C & L-RMU  & \bf{0.29} & 0.32 & 0.43 & 0.42 & --- & --- \\
 C & RMU-Split  & 0.54 & 0.46 & 0.54 & 0.47 & --- & --- \\
 C & L-RMU-Split  & 0.29 & \bf{0.31} & \bf{0.42} & \bf{0.38} & --- & --- \\
\midrule
 A, B & RMU  & --- & --- & --- & --- & 0.61 & 0.58 \\
 A, B & L-RMU  & --- & --- & --- & --- & \bf{0.51} & \bf{0.52} \\
 A, B & RMU-Split  & --- & --- & --- & --- & 0.59 & 0.54 \\
 A, B & L-RMU-Split  & --- & --- & --- & --- & 0.56 & 0.53 \\
\midrule
 A, C & RMU  & --- & --- & 0.61 & 0.57 & --- & --- \\
 A, C & L-RMU  & --- & --- & 0.52 & 0.45 & --- & --- \\
 A, C & RMU-Split  & --- & --- & 0.60 & 0.51 & --- & --- \\
 A, C & L-RMU-Split  & --- & --- & \bf{0.51} & \bf{0.42} & --- & --- \\
\midrule
 B, C & RMU  & 0.62 & 0.59 & --- & --- & --- & --- \\
 B, C & L-RMU  & 0.43 & \bf{0.32} & --- & --- & --- & --- \\
 B, C & RMU-Split  & 0.59 & 0.50 & --- & --- & --- & --- \\
 B, C & L-RMU-Split  & \bf{0.40} & 0.34 & --- & --- & --- & --- \\
\bottomrule
\end{tabular}
\end{center}
\end{table}

\pagebreak

\section{Relearning recovery rates}

We define the \textbf{recovery rate}. Consider two resulting models, $P$ and $Q$, after unlearning. We aim to compare the percentage of information recovered by adversarial relearning on each model. We define the recovery rate as:
\begin{equation*}
\text{Recovery Rate} = \frac{\text{P Accuracy After Relearning} - \text{P Accuracy After Unlearning}}{\text{Q Accuracy After Relearning} - \text{Q Accuracy After Unlearning}}.
\end{equation*}
This metric quantifies how much more vulnerable model $P$ is to relearning compared to model $Q$, with a lower recovery rate indicating better unlearning robustness.

To ensure fair comparisons, we impose a minimum unlearning accuracy threshold of $0.25$ (random guessing for MCQ) in cases where the model is highly unlearned. This prevents artificially inflating the recovery rate when the model has already reached the theoretical lower bound of performance.

When computing recovery rates, we compare how much performance is regained between two unlearning algorithms, effectively normalizing for the choice of base method. We specify each algorithm and its base algorithm as follows:
\begin{itemize}
    % \item RMU-LAT: RMU.
    \item L-RMU: RMU.
    \item L-RMU-Split: RMU-Split.
    \item L-SimNPO: SimNPO.
\end{itemize}

\subsection{WMDP 2 folds}
\begin{table}[H]
\caption{Recovery rate across methods for WMDP 2 folds.}
\label{tab:wmdp_2_recover_rate}
\begin{center}
\begin{tabular}{clcccc}
\toprule
\multirow{2}{*}{\bf Relearn} & \multirow{2}{*}{\bf Method} &\multicolumn{2}{c}{\bf A $\downarrow$} & \multicolumn{2}{c}{\bf B $\downarrow$} \\
\cmidrule(lr){3-4} \cmidrule(lr){5-6} 
 & & MCQ & Corpus & MCQ & Corpus \\
\midrule
 A & L-RMU  & --- & --- & 0.95 & 1.35 \\
 A & L-RMU-Split  & --- & --- & \bf{0.92} & \bf{1.07} \\
\midrule
 B & L-RMU  & \bf{0.26} & \bf{0.66} & --- & --- \\
 B & L-RMU-Split  & 0.43 & 0.82 & --- & --- \\
\bottomrule
\end{tabular}
\end{center}
\end{table}

\subsection{WMDP 3 folds}
\begin{table}[H]
\caption{Recovery rate across methods for WMDP 3 folds.}
\label{tab:wmdp_3_recover_rate}
\begin{center}
\begin{tabular}{clcccccc}
\toprule
\multirow{2}{*}{\bf Relearn} & \multirow{2}{*}{\bf Method} &\multicolumn{2}{c}{\bf A $\downarrow$} & \multicolumn{2}{c}{\bf B $\downarrow$} & \multicolumn{2}{c}{\bf C $\downarrow$} \\
\cmidrule(lr){3-4} \cmidrule(lr){5-6} \cmidrule(lr){7-8} 
 & & MCQ & Corpus & MCQ & Corpus & MCQ & Corpus \\
\midrule
 % A & RMU-LAT  & --- & --- & 0.88 & 0.96 & 1.76 & 1.75 \\
 A & L-RMU  & --- & --- & 0.91 & 1.21 & 1.33 & 1.25 \\
 A & L-RMU-Split  & --- & --- & 0.84 & 1.04 & 1.77 & 0.96 \\
 A & L-SimNPO  & --- & --- & \bf{0.40} & \bf{0.17} & \bf{0.40} & \bf{0.30} \\
\midrule
 % B & RMU-LAT  & 1.07 & 1.10 & --- & --- & 1.40 & 1.51 \\
 B & L-RMU  & 0.42 & 0.88 & --- & --- & 1.29 & 1.44 \\
 B & L-RMU-Split  & \bf{0.26} & \bf{0.64} & --- & --- & 1.54 & 0.95 \\
 B & L-SimNPO  & 0.58 & 0.79 & --- & --- & \bf{0.31} & \bf{0.38} \\
\midrule
 % C & RMU-LAT  & 0.85 & 1.07 & 1.19 & 0.92 & --- & --- \\
 C & L-RMU  & 0.06 & 0.48 & 0.62 & 0.88 & --- & --- \\
 C & L-RMU-Split  & \bf{0.00} & \bf{0.15} & \bf{0.15} & \bf{0.43} & --- & --- \\
 C & L-SimNPO  & 0.63 & 0.76 & 0.58 & 0.51 & --- & --- \\
\midrule
 % A, B & RMU-LAT  & --- & --- & --- & --- & 1.63 & 1.68 \\
 A, B & L-RMU  & --- & --- & --- & --- & 1.05 & 1.26 \\
 A, B & L-RMU-Split  & --- & --- & --- & --- & 1.40 & 0.91 \\
 A, B & L-SimNPO  & --- & --- & --- & --- & \bf{0.39} & \bf{0.33} \\
\midrule
 % A, C & RMU-LAT  & --- & --- & 0.83 & 1.02 & --- & --- \\
 A, C & L-RMU  & --- & --- & 0.71 & 1.14 & --- & --- \\
 A, C & L-RMU-Split  & --- & --- & 0.51 & 1.00 & --- & --- \\
 A, C & L-SimNPO  & --- & --- & \bf{0.50} & \bf{0.42} & --- & --- \\
\midrule
 % B, C & RMU-LAT  & 0.87 & 1.10 & --- & --- & --- & --- \\
 B, C & L-RMU  & \bf{0.58} & 0.78 & --- & --- & --- & --- \\
 B, C & L-RMU-Split  & 0.73 & \bf{0.56} & --- & --- & --- & --- \\
 B, C & L-SimNPO  & 0.75 & 0.74 & --- & --- & --- & --- \\
\bottomrule
\end{tabular}
\end{center}
\end{table}

\pagebreak
\subsection{WMDP 4 folds}
\begin{longtable}{clcccccccc}
\caption{Recovery rate across methods for WMDP 4 folds.}\\
\label{tab:wmdp_4_recover_rate}\\
\toprule
\multirow{2}{*}{\bf Relearn} & \multirow{2}{*}{\bf Method} &\multicolumn{2}{c}{\bf A $\downarrow$} & \multicolumn{2}{c}{\bf B $\downarrow$} & \multicolumn{2}{c}{\bf C $\downarrow$} & \multicolumn{2}{c}{\bf D $\downarrow$} \\
\cmidrule(lr){3-4} \cmidrule(lr){5-6} \cmidrule(lr){7-8} \cmidrule(lr){9-10} 
 & & MCQ & Corpus & MCQ & Corpus & MCQ & Corpus & MCQ & Corpus \\
\midrule
\endfirsthead
\multicolumn{10}{c}{\textit{Continued from previous page}} \\
\\[1mm] % Adjust the space as needed
\toprule
\multirow{2}{*}{\bf Relearn} & \multirow{2}{*}{\bf Method} &\multicolumn{2}{c}{\bf A $\downarrow$} & \multicolumn{2}{c}{\bf B $\downarrow$} & \multicolumn{2}{c}{\bf C $\downarrow$} & \multicolumn{2}{c}{\bf D $\downarrow$} \\
\cmidrule(lr){3-4} \cmidrule(lr){5-6} \cmidrule(lr){7-8} \cmidrule(lr){9-10} 
 & & MCQ & Corpus & MCQ & Corpus & MCQ & Corpus & MCQ & Corpus \\
\midrule
\endhead
\\[1mm] % Adjust the space as needed
\multicolumn{10}{c}{\textit{Continued on next page}} \\
\endfoot
\bottomrule
\endlastfoot
 A & L-RMU  & --- & --- & 0.93 & 1.24 & 1.19 & 1.37 & \bf{0.62} & 1.35 \\
 A & L-RMU-Split  & --- & --- & \bf{0.78} & \bf{1.22} & \bf{0.96} & \bf{1.08} & 0.68 & \bf{0.84} \\
\midrule
 B & L-RMU  & \bf{0.21} & 1.18 & --- & --- & \bf{0.71} & 1.71 & \bf{0.68} & 1.63 \\
 B & L-RMU-Split  & 0.56 & \bf{0.97} & --- & --- & 0.83 & \bf{1.05} & 1.13 & \bf{0.86} \\
\midrule
 C & L-RMU  & 0.43 & \bf{0.86} & \bf{0.77} & 1.10 & --- & --- & 0.75 & 1.59 \\
 C & L-RMU-Split  & \bf{0.26} & 0.97 & 1.05 & \bf{1.10} & --- & --- & \bf{0.74} & \bf{0.89} \\
\midrule
 D & L-RMU  & \bf{0.00} & \bf{0.37} & \bf{0.10} & \bf{0.85} & 1.00 & 1.63 & --- & --- \\
 D & L-RMU-Split  & \bf{0.00} & 0.61 & 0.56 & 0.92 & \bf{0.56} & \bf{0.68} & --- & --- \\
\midrule
 A, B & L-RMU  & --- & --- & --- & --- & 1.40 & 1.31 & \bf{0.71} & 1.00 \\
 A, B & L-RMU-Split  & --- & --- & --- & --- & \bf{0.96} & \bf{1.05} & 1.25 & \bf{0.81} \\
\midrule
 A, C & L-RMU  & --- & --- & \bf{1.30} & \bf{1.07} & --- & --- & \bf{0.64} & 1.22 \\
 A, C & L-RMU-Split  & --- & --- & 1.33 & 1.18 & --- & --- & 1.04 & \bf{0.80} \\
\midrule
 A, D & L-RMU  & --- & --- & \bf{0.65} & \bf{1.05} & 1.18 & 1.27 & --- & --- \\
 A, D & L-RMU-Split  & --- & --- & 0.90 & 1.09 & \bf{0.84} & \bf{0.98} & --- & --- \\
\midrule
 B, C & L-RMU  & \bf{0.26} & 1.00 & --- & --- & --- & --- & \bf{0.69} & 1.07 \\
 B, C & L-RMU-Split  & 0.68 & \bf{0.88} & --- & --- & --- & --- & 1.47 & \bf{0.80} \\
\midrule
 B, D & L-RMU  & \bf{0.48} & \bf{0.85} & --- & --- & 1.13 & 1.29 & --- & --- \\
 B, D & L-RMU-Split  & 0.52 & 0.90 & --- & --- & \bf{0.93} & \bf{1.07} & --- & --- \\
\midrule
 C, D & L-RMU  & \bf{0.11} & \bf{0.77} & \bf{0.66} & \bf{0.95} & --- & --- & --- & --- \\
 C, D & L-RMU-Split  & 0.55 & 0.85 & 1.24 & 1.11 & --- & --- & --- & --- \\
\midrule
 A, B, C & L-RMU  & --- & --- & --- & --- & --- & --- & \bf{0.94} & 1.02 \\
 A, B, C & L-RMU-Split  & --- & --- & --- & --- & --- & --- & 1.48 & \bf{0.80} \\
\midrule
 A, B, D & L-RMU  & --- & --- & --- & --- & \bf{1.10} & 1.31 & --- & --- \\
 A, B, D & L-RMU-Split  & --- & --- & --- & --- & 1.12 & \bf{1.00} & --- & --- \\
\midrule
 A, C, D & L-RMU  & --- & --- & \bf{0.67} & \bf{1.05} & --- & --- & --- & --- \\
 A, C, D & L-RMU-Split  & --- & --- & 0.77 & 1.09 & --- & --- & --- & --- \\
\midrule
 B, C, D & L-RMU  & \bf{0.30} & 0.87 & --- & --- & --- & --- & --- & --- \\
 B, C, D & L-RMU-Split  & 0.77 & \bf{0.80} & --- & --- & --- & --- & --- & --- \\
\end{longtable}

\pagebreak
\subsection{MMLU 3 folds}
\begin{table}[H]
\caption{Recovery rate across methods for MMLU 3 folds.}
\label{tab:mmlu_3_recover_rate}
\begin{center}
\begin{tabular}{clcccccc}
\toprule
\multirow{2}{*}{\bf Relearn} & \multirow{2}{*}{\bf Method} &\multicolumn{2}{c}{\bf A $\downarrow$} & \multicolumn{2}{c}{\bf B $\downarrow$} & \multicolumn{2}{c}{\bf C $\downarrow$} \\
\cmidrule(lr){3-4} \cmidrule(lr){5-6} \cmidrule(lr){7-8} 
 & & MCQ & Corpus & MCQ & Corpus & MCQ & Corpus \\
\midrule
 A & L-RMU  & --- & --- & \bf{0.61} & \bf{0.89} & 1.04 & \bf{0.87} \\
 A & L-RMU-Split  & --- & --- & 0.63 & 0.96 & \bf{1.02} & 1.01 \\
\midrule
 B & L-RMU  & \bf{0.20} & \bf{0.79} & --- & --- & 1.20 & \bf{0.93} \\
 B & L-RMU-Split  & 0.53 & 0.96 & --- & --- & \bf{0.92} & 0.98 \\
\midrule
 C & L-RMU  & \bf{0.33} & \bf{0.38} & 0.44 & \bf{0.51} & --- & --- \\
 C & L-RMU-Split  & 0.38 & 0.59 & \bf{0.40} & 0.71 & --- & --- \\
\midrule
 A, B & L-RMU  & --- & --- & --- & --- & 1.05 & \bf{0.89} \\
 A, B & L-RMU-Split  & --- & --- & --- & --- & \bf{0.70} & 0.93 \\
\midrule
 A, C & L-RMU  & --- & --- & \bf{0.59} & \bf{0.93} & --- & --- \\
 A, C & L-RMU-Split  & --- & --- & 0.72 & 0.94 & --- & --- \\
\midrule
 B, C & L-RMU  & \bf{0.30} & \bf{0.82} & --- & --- & --- & --- \\
 B, C & L-RMU-Split  & 0.47 & 0.84 & --- & --- & --- & --- \\
\bottomrule
\end{tabular}
\end{center}
\end{table}

\subsection{Years 3 folds}
\begin{table}[H]
\caption{Recovery rate across methods for Years 3 folds.}
\label{tab:years_3_recover_rate}
\begin{center}
\begin{tabular}{clcccccc}
\toprule
\multirow{2}{*}{\bf Relearn} & \multirow{2}{*}{\bf Method} &\multicolumn{2}{c}{\bf A $\downarrow$} & \multicolumn{2}{c}{\bf B $\downarrow$} & \multicolumn{2}{c}{\bf C $\downarrow$} \\
\cmidrule(lr){3-4} \cmidrule(lr){5-6} \cmidrule(lr){7-8} 
 & & MCQ & Corpus & MCQ & Corpus & MCQ & Corpus \\
\midrule
 A & L-RMU  & --- & --- & \bf{0.65} & \bf{0.53} & \bf{0.77} & \bf{0.79} \\
 A & L-RMU-Split  & --- & --- & 0.67 & 0.56 & 0.84 & 1.01 \\
\midrule
 B & L-RMU  & 0.42 & \bf{0.17} & --- & --- & \bf{0.81} & \bf{0.70} \\
 B & L-RMU-Split  & \bf{0.32} & 0.27 & --- & --- & 1.04 & 0.90 \\
\midrule
 C & L-RMU  & \bf{0.00} & \bf{0.11} & \bf{0.46} & \bf{0.44} & --- & --- \\
 C & L-RMU-Split  & 0.02 & 0.17 & 0.67 & 0.69 & --- & --- \\
\midrule
 A, B & L-RMU  & --- & --- & --- & --- & \bf{0.66} & \bf{0.78} \\
 A, B & L-RMU-Split  & --- & --- & --- & --- & 0.86 & 0.89 \\
\midrule
 A, C & L-RMU  & --- & --- & \bf{0.67} & \bf{0.55} & --- & --- \\
 A, C & L-RMU-Split  & --- & --- & 0.84 & 0.76 & --- & --- \\
\midrule
 B, C & L-RMU  & 0.46 & \bf{0.13} & --- & --- & --- & --- \\
 B, C & L-RMU-Split  & \bf{0.41} & 0.29 & --- & --- & --- & --- \\
\bottomrule
\end{tabular}
\end{center}
\end{table}

\pagebreak

\section{Hyperparameters}
\label{appendix:hyperparameters}

We set the forgetting threshold to $0.35$, motivated by the following statistical reasoning. 

For multiple-choice questions (MCQ) with $4$ choices, assuming random guessing, the expected accuracy is $0.25$. Each dataset contains a total of $735$ questions, and the smallest fold we consider consists of $\frac{735}{4}$. By applying the central limit theorem, the expected final accuracy follows approximately a normal distribution:
\begin{equation*}
\mathcal{N} \left(\frac{1}{4}, \sqrt{\frac{1}{4} \cdot \frac{3}{4} \cdot \frac{4}{735}} \right) \approx \mathcal{N} \left(0.25, 0.032\right).
\end{equation*}
A three-standard-deviation event corresponds to:
\begin{equation*}
3 \times 0.032 + 0.25 \leq 0.35.
\end{equation*}
Since we do not reject the null hypothesis of random guessing if accuracy remains within three standard deviations of $0.25$, we set the forgetting threshold to approximately $0.35$. Note that this bound does get tighter if we consider bigger folds, but in practice, we find that this threshold does not significantly impact results, so we standardize it across all instances of LU.

For all models, we use Zephyr-$7B$-$\beta$ \citep{tunstall2023zephyrdirectdistillationlm}.

\subsection{RMU hyperparameters}

WMDP: official model checkpoint from \citep{li2024wmdpbenchmarkmeasuringreducing}.
MMLU:
\begin{itemize}
    \item Activation layer: $7$.
    \item Layers fine-tuned: $5,6,7$.
    \item Magnitude: $10$.
    \item Forget coefficient: $2.00$.
    \item Retain coefficient: $16.00$.
    \item Learning rate: $5\times 10^{-5}$.
    \item Batch size: $4$.
\end{itemize}

Years:
\begin{itemize}
    \item Activation layer: $7$.
    \item Layers fine-tuned: $5,6,7$.
    \item Magnitude: $15$.
    \item Forget coefficient: $0.25$.
    \item Retain coefficient: $1.00$.
    \item Learning rate: $5\times 10^{-5}$.
    \item Batch size: $4$.
\end{itemize}

\subsection{RMU-Split hyperparameters} 

WMDP $2$ folds: 
\begin{itemize}
\item Activation layer: $7$.
\item Layers fine-tuned: $5, 6, 7$.
\item Magnitude: $10$.
\item Forget coefficients: $1.00, 1.00$.
\item Retain coefficient $32$.
\item Learning rate: $1 \times 10^{-5}$.
\item Batch size: $4$.
\end{itemize}
WMDP $3$ folds: 
\begin{itemize}
\item Activation layer: $7$.
\item Layers fine-tuned: $5, 6, 7$.
\item Magnitude: $10$.
\item Forget coefficients: $1.00, 1.00, 1.00$.
\item Retain coefficient $16$.
\item Learning rate: $1 \times 10^{-5}$.
\item Batch size: $4$.
\end{itemize}
WMDP $4$ folds: 
\begin{itemize}
\item Activation layer: $7$.
\item Layers fine-tuned: $5, 6, 7$.
\item Magnitude: $10$.
\item Forget coefficients: $1.00, 1.00, 1.00, 1.00$.
\item Retain coefficient $32$.
\item Learning rate: $1 \times 10^{-5}$.
\item Batch size: $4$.
\end{itemize}
MMLU $3$ folds: 
\begin{itemize}
\item Activation layer: $7$.
\item Layers fine-tuned: $5, 6, 7$.
\item Magnitude: $10$.
\item Forget coefficients: $2.00, 2.00, 2.00$.
\item Retain coefficient $24$.
\item MMLU retain coefficient: $12.0$.
\item Learning rate: $1 \times 10^{-5}$.
\item Batch size: $8$.
\end{itemize}
Years $3$ folds: 
\begin{itemize}
\item Activation layer: $7$.
\item Layers fine-tuned: $5, 6, 7$.
\item Magnitude: $10$.
\item Forget coefficients: $1.00, 1.00, 1.00$.
\item Retain coefficient $32$.
\item Learning rate: $1 \times 10^{-5}$.
\item Batch size: $4$.
\end{itemize}

% \subsection{RMU-LAT hyperparameters}
% WMDP: official model checkpoint from \citep{sheshadri2025latent}.

\subsection{L-RMU hyperparameters}
WMDP $2$ folds: 
\begin{itemize}
\item Stage $1$:
\begin{itemize}
\item Activation layer: $7$.
\item Layers fine-tuned: $5, 6, 7$.
\item Magnitude: $6.5$.
\item Forget coefficients: $0.39, 0.00$.
\item Retain coefficients: $0.00, 13.52$.
\item Retain set coefficient: $1$.
\item Learning rate: $1 \times 10^{-5}$.
\item Batch size: $4$.
\end{itemize}
\item Stage $2$:
\begin{itemize}
\item Activation layer: $7$.
\item Layers fine-tuned: $5, 6, 7$.
\item Magnitude: $6.5$.
\item Forget coefficients: $0.10, 1.00$.
\item Retain coefficients: $0.00, 0.00$.
\item Retain set coefficient: $24$.
\item Learning rate: $1 \times 10^{-5}$.
\item Batch size: $8$.
\end{itemize}
\end{itemize}
WMDP $3$ folds: 
\begin{itemize}
\item Stage $1$:
\begin{itemize}
\item Activation layer: $7$.
\item Layers fine-tuned: $5, 6, 7$.
\item Magnitude: $6.5$.
\item Forget coefficients: $0.39, 0.00, 0.00$.
\item Retain coefficients: $0.00, 6.76, 6.76$.
\item Retain set coefficient: $1$.
\item Learning rate: $1 \times 10^{-5}$.
\item Batch size: $4$.
\end{itemize}
\item Stage $2$:
\begin{itemize}
\item Activation layer: $7$.
\item Layers fine-tuned: $5, 6, 7$.
\item Magnitude: $6.5$.
\item Forget coefficients: $0.05, 0.50, 0.00$.
\item Retain coefficients: $0.00, 0.00, 8.00$.
\item Retain set coefficient: $16$.
\item Learning rate: $1 \times 10^{-5}$.
\item Batch size: $8$.
\end{itemize}
\item Stage $3$:
\begin{itemize}
\item Activation layer: $7$.
\item Layers fine-tuned: $5, 6, 7$.
\item Magnitude: $6.5$.
\item Forget coefficients: $0.00, 0.03, 0.33$.
\item Retain coefficients: $0.00, 0.00, 0.00$.
\item Retain set coefficient: $24$.
\item Learning rate: $1 \times 10^{-5}$.
\item Batch size: $8$.
\end{itemize}
\end{itemize}
WMDP $4$ folds: 
\begin{itemize}
\item Stage $1$:
\begin{itemize}
\item Activation layer: $7$.
\item Layers fine-tuned: $5, 6, 7$.
\item Magnitude: $6.5$.
\item Forget coefficients: $0.39, 0.00, 0.00, 0.00$.
\item Retain coefficients: $0.00, 10.67, 10.67, 10.67$.
\item Retain set coefficient: $1$.
\item Learning rate: $1 \times 10^{-5}$.
\item Batch size: $4$.
\end{itemize}
\item Stage $2$:
\begin{itemize}
\item Activation layer: $7$.
\item Layers fine-tuned: $5, 6, 7$.
\item Magnitude: $6.5$.
\item Forget coefficients: $0.10, 1.00, 0.00, 0.00$.
\item Retain coefficients: $0.00, 0.00, 4.00, 4.00$.
\item Retain set coefficient: $16$.
\item Learning rate: $1 \times 10^{-5}$.
\item Batch size: $8$.
\end{itemize}
\item Stage $3$:
\begin{itemize}
\item Activation layer: $7$.
\item Layers fine-tuned: $5, 6, 7$.
\item Magnitude: $6.5$.
\item Forget coefficients: $0.01, 0.07, 0.67, 0.00$.
\item Retain coefficients: $0.00, 0.00, 0.00, 8.00$.
\item Retain set coefficient: $16$.
\item Learning rate: $1 \times 10^{-5}$.
\item Batch size: $8$.
\end{itemize}
\item Stage $4$:
\begin{itemize}
\item Activation layer: $7$.
\item Layers fine-tuned: $5, 6, 7$.
\item Magnitude: $6.5$.
\item Forget coefficients: $0.01, 0.03, 0.15, 0.75$.
\item Retain coefficients: $0.00, 0.00, 0.00, 0.00$.
\item Retain set coefficient: $32$.
\item Learning rate: $1 \times 10^{-5}$.
\item Batch size: $8$.
\end{itemize}
\end{itemize}
MMLU $3$ folds: 
\begin{itemize}
\item Stage $1$:
\begin{itemize}
\item Activation layer: $7$.
\item Layers fine-tuned: $5, 6, 7$.
\item Magnitude: $6.5$.
\item Forget coefficients: $2.00, 0.00, 0.00$.
\item Retain coefficients: $0.00, 8.00, 8.00$.
\item Retain set coefficient: $2$.
\item Learning rate: $1 \times 10^{-5}$.
\item Batch size: $4$.
\end{itemize}
\item Stage $2$:
\begin{itemize}
\item Activation layer: $7$.
\item Layers fine-tuned: $5, 6, 7$.
\item Magnitude: $6.5$.
\item Forget coefficients: $0.10, 2.00, 0.00$.
\item Retain coefficients: $0.00, 0.00, 4.00$.
\item Retain set coefficient: $32$.
\item MMLU retain coefficient: $8.0$.
\item Learning rate: $1 \times 10^{-5}$.
\item Batch size: $8$.
\end{itemize}
\item Stage $3$:
\begin{itemize}
\item Activation layer: $7$.
\item Layers fine-tuned: $5, 6, 7$.
\item Magnitude: $6.5$.
\item Forget coefficients: $0.01, 0.13, 2.67$.
\item Retain coefficients: $0.00, 0.00, 0.00$.
\item Retain set coefficient: $36$.
\item MMLU retain coefficient: $18.0$.
\item Learning rate: $1 \times 10^{-5}$.
\item Batch size: $8$.
\end{itemize}
\end{itemize}
Years $3$ folds: 
\begin{itemize}
\item Stage $1$:
\begin{itemize}
\item Activation layer: $7$.
\item Layers fine-tuned: $5, 6, 7$.
\item Magnitude: $6.5$.
\item Forget coefficients: $4.00, 0.00, 0.00$.
\item Retain coefficients: $0.00, 16.00, 16.00$.
\item Retain set coefficient: $2$.
\item Learning rate: $1 \times 10^{-5}$.
\item Batch size: $4$.
\end{itemize}
\item Stage $2$:
\begin{itemize}
\item Activation layer: $7$.
\item Layers fine-tuned: $5, 6, 7$.
\item Magnitude: $6.5$.
\item Forget coefficients: $2.25, 22.50, 0.00$.
\item Retain coefficients: $0.00, 0.00, 16.00$.
\item Retain set coefficient: $16$.
\item Learning rate: $1 \times 10^{-5}$.
\item Batch size: $8$.
\end{itemize}
\item Stage $3$:
\begin{itemize}
\item Activation layer: $7$.
\item Layers fine-tuned: $5, 6, 7$.
\item Magnitude: $6.5$.
\item Forget coefficients: $0.15, 1.50, 15.00$.
\item Retain coefficients: $0.00, 0.00, 0.00$.
\item Retain set coefficient: $32$.
\item Learning rate: $1 \times 10^{-5}$.
\item Batch size: $8$.
\end{itemize}
\end{itemize}

\subsection{L-RMU-Split hyperparameters}

WMDP $2$ folds: 
\begin{itemize}
\item Stage $1$:
\begin{itemize}
\item Activation layer: $7$.
\item Layers fine-tuned: $5, 6, 7$.
\item Magnitude: $6.5$.
\item Forget coefficients: $0.39, 0.00$.
\item Retain coefficients: $0.00, 13.52$.
\item Retain set coefficient: $1$.
\item Learning rate: $1 \times 10^{-5}$.
\item Batch size: $4$.
\end{itemize}
\item Stage $2$:
\begin{itemize}
\item Activation layer: $7$.
\item Layers fine-tuned: $5, 6, 7$.
\item Magnitude: $6.5$.
\item Forget coefficients: $0.10, 0.20$.
\item Retain coefficients: $0.00, 0.00$.
\item Retain set coefficient: $14.51609$.
\item Learning rate: $1 \times 10^{-5}$.
\item Batch size: $4$.
\end{itemize}
\end{itemize}
WMDP $3$ folds: 
\begin{itemize}
\item Stage $1$:
\begin{itemize}
\item Activation layer: $7$.
\item Layers fine-tuned: $5, 6, 7$.
\item Magnitude: $6.5$.
\item Forget coefficients: $0.39, 0.00, 0.00$.
\item Retain coefficients: $0.00, 6.76, 6.76$.
\item Retain set coefficient: $1$.
\item Learning rate: $1 \times 10^{-5}$.
\item Batch size: $4$.
\end{itemize}
\item Stage $2$:
\begin{itemize}
\item Activation layer: $7$.
\item Layers fine-tuned: $5, 6, 7$.
\item Magnitude: $6.5$.
\item Forget coefficients: $1.00, 4.00, 0.00$.
\item Retain coefficients: $0.00, 0.00, 13.52$.
\item Retain set coefficient: $32$.
\item Learning rate: $3 \times 10^{-6}$.
\item Batch size: $4$.
\end{itemize}
\item Stage $3$:
\begin{itemize}
\item Activation layer: $7$.
\item Layers fine-tuned: $5, 6, 7$.
\item Magnitude: $6.5$.
\item Forget coefficients: $0.33, 1.33, 5.33$.
\item Retain coefficients: $0.00, 0.00, 0.00$.
\item Retain set coefficient: $45.51609$.
\item Learning rate: $3 \times 10^{-6}$.
\item Batch size: $12$.
\end{itemize}
\end{itemize}
WMDP $4$ folds: 
\begin{itemize}
\item Stage $1$:
\begin{itemize}
\item Activation layer: $7$.
\item Layers fine-tuned: $5, 6, 7$.
\item Magnitude: $6.5$.
\item Forget coefficients: $0.39, 0.00, 0.00, 0.00$.
\item Retain coefficients: $0.00, 10.67, 10.67, 10.67$.
\item Retain set coefficient: $1$.
\item Learning rate: $1 \times 10^{-5}$.
\item Batch size: $4$.
\end{itemize}
\item Stage $2$:
\begin{itemize}
\item Activation layer: $7$.
\item Layers fine-tuned: $5, 6, 7$.
\item Magnitude: $6.5$.
\item Forget coefficients: $0.10, 0.20, 0.00, 0.00$.
\item Retain coefficients: $0.00, 0.00, 16.00, 16.00$.
\item Retain set coefficient: $8$.
\item Learning rate: $1 \times 10^{-5}$.
\item Batch size: $4$.
\end{itemize}
\item Stage $3$:
\begin{itemize}
\item Activation layer: $7$.
\item Layers fine-tuned: $5, 6, 7$.
\item Magnitude: $6.5$.
\item Forget coefficients: $0.03, 0.07, 0.13, 0.00$.
\item Retain coefficients: $0.00, 0.00, 0.00, 32.00$.
\item Retain set coefficient: $16$.
\item Learning rate: $1 \times 10^{-5}$.
\item Batch size: $4$.
\end{itemize}
\item Stage $4$:
\begin{itemize}
\item Activation layer: $7$.
\item Layers fine-tuned: $5, 6, 7$.
\item Magnitude: $6.5$.
\item Forget coefficients: $0.03, 0.06, 0.12, 0.25$.
\item Retain coefficients: $0.00, 0.00, 0.00, 0.00$.
\item Retain set coefficient: $32$.
\item Learning rate: $1 \times 10^{-5}$.
\item Batch size: $4$.
\end{itemize}
\end{itemize}
MMLU $3$ folds: 
\begin{itemize}
\item Stage $1$:
\begin{itemize}
\item Activation layer: $7$.
\item Layers fine-tuned: $5, 6, 7$.
\item Magnitude: $6.5$.
\item Forget coefficients: $2.00, 0.00, 0.00$.
\item Retain coefficients: $0.00, 8.00, 8.00$.
\item Retain set coefficient: $2$.
\item Learning rate: $1 \times 10^{-5}$.
\item Batch size: $4$.
\end{itemize}
\item Stage $2$:
\begin{itemize}
\item Activation layer: $7$.
\item Layers fine-tuned: $5, 6, 7$.
\item Magnitude: $6.5$.
\item Forget coefficients: $0.10, 2.00, 0.00$.
\item Retain coefficients: $0.00, 0.00, 8.00$.
\item Retain set coefficient: $32$.
\item MMLU retain coefficient: $8.0$.
\item Learning rate: $1 \times 10^{-5}$.
\item Batch size: $4$.
\end{itemize}
\item Stage $3$:
\begin{itemize}
\item Activation layer: $7$.
\item Layers fine-tuned: $5, 6, 7$.
\item Magnitude: $10$.
\item Forget coefficients: $0.00, 0.07, 1.33$.
\item Retain coefficients: $0.00, 0.00, 0.00$.
\item Retain set coefficient: $40$.
\item MMLU retain coefficient: $10.0$.
\item Learning rate: $1 \times 10^{-5}$.
\item Batch size: $4$.
\end{itemize}
\end{itemize}
Years $3$ folds: 
\begin{itemize}
\item Stage $1$:
\begin{itemize}
\item Activation layer: $7$.
\item Layers fine-tuned: $5, 6, 7$.
\item Magnitude: $6.5$.
\item Forget coefficients: $4.00, 0.00, 0.00$.
\item Retain coefficients: $0.00, 16.00, 16.00$.
\item Retain set coefficient: $2$.
\item Learning rate: $1 \times 10^{-5}$.
\item Batch size: $4$.
\end{itemize}
\item Stage $2$:
\begin{itemize}
\item Activation layer: $7$.
\item Layers fine-tuned: $5, 6, 7$.
\item Magnitude: $12$.
\item Forget coefficients: $1.20, 4.00, 0.00$.
\item Retain coefficients: $0.00, 0.00, 32.00$.
\item Retain set coefficient: $2$.
\item Learning rate: $1 \times 10^{-5}$.
\item Batch size: $4$.
\end{itemize}
\item Stage $3$:
\begin{itemize}
\item Activation layer: $7$.
\item Layers fine-tuned: $5, 6, 7$.
\item Magnitude: $12$.
\item Forget coefficients: $0.17, 0.67, 2.67$.
\item Retain coefficients: $0.00, 0.00, 0.00$.
\item Retain set coefficient: $36$.
\item Learning rate: $1 \times 10^{-5}$.
\item Batch size: $4$.
\end{itemize}
\end{itemize}

\subsection{SimNPO hyperparameters}

WMDP: 
\begin{itemize}
\item Beta: $0.1$.
\item Forget coefficients: $8.00$.
\item Retain coefficients: $1.00$.
\item Learning rate: $4 \times 10^{-6}$.
\item Batch size: $4$.
\end{itemize}

We reran SimNPO with our own implementation as the checkpoint online did not sufficiently unlearn the data on our folds. 

\subsection{L-SimNPO hyperparameters}

WMDP $3$ folds: 
\begin{itemize}
\item Stage $1$:
\begin{itemize}
\item Beta: $0.1$.
\item Forget coefficients: $3.00, 0.00, 0.00$.
\item Retain coefficients: $0.00, 1.00, 1.00$.
\item Retain set coefficient: $4$.
\item Learning rate: $4 \times 10^{-6}$.
\item Batch size: $4$.
\end{itemize}
\item Stage $2$:
\begin{itemize}
\item Beta: $0.1$.
\item Forget coefficients: $0.60, 3.00, 0.00$.
\item Retain coefficients: $0.00, 0.00, 1.00$.
\item Retain set coefficient: $6$.
\item Learning rate: $4 \times 10^{-6}$.
\item Batch size: $4$.
\end{itemize}
\item Stage $3$:
\begin{itemize}
\item Beta: $0.1$.
\item Forget coefficients: $5.00, 5.00, 5.00$.
\item Retain coefficients: $0.00, 0.00, 0.00$.
\item Retain set coefficient: $6$.
\item Learning rate: $4 \times 10^{-6}$.
\item Batch size: $4$.
\end{itemize}
\end{itemize}

\end{document}